\documentclass[final,12pt]{elsarticle}

\usepackage{hyperref}
\usepackage[caption=false]{subfig}
\usepackage[normalem]{ulem}
\graphicspath{{./fig/}}
\DeclareGraphicsExtensions{.pdf,.jpeg,.png}
\usepackage{relsize}
\usepackage[cmex10]{amsmath}

\usepackage[absolute]{textpos}
\usepackage{color}

\journal{Journal of Medical Image Analysis}

\biboptions{authoryear}

\begin{document}

\begin{textblock}{14}(1,1)
\noindent \color{blue} Preprint of the accepted manuscript in the Journal of Medical Image Computing. \textcopyright 2016. \\
 DOI: 10.1016/j.media.2016.03.002.\\ This manuscript version is made available under the CC-BY 4.0 license \\ \url{https://creativecommons.org/licenses/by/4.0/} 
\end{textblock}

\begin{frontmatter}

\title{Pieces-of-parts for supervoxel segmentation with global context: Application to DCE-MRI tumour delineation} 

\author[ibme]{Benjamin Irving\corref{mycorrespondingauthor}}
\cortext[mycorrespondingauthor]{Corresponding author}
\ead{benjamin.irving@eng.ox.ac.uk}

\author[radiology]{James M. Franklin}
\author[ibme]{Bart\l omiej W. Papie\.z}
\author[radiology]{Ewan M. Anderson}
\author[oncology]{Ricky A. Sharma}
\author[radiology]{Fergus V. Gleeson}
\author[oncology]{Sir Michael Brady}
\author[ibme,kings]{Julia A. Schnabel}

\address[ibme]{Institute of Biomedical Engineering, Department of Engineering Science, Old Road Campus Research Building, University of Oxford, Headington, Oxford, OX3 7DQ, UK}
\address[radiology]{Department of Radiology, Oxford University Hospitals NHS Foundation Trust, Churchill Hospital, Oxford, OX3 7LE, UK}
\address[oncology]{Department of Oncology, Old Road Campus Research Building, University of Oxford, Headington, Oxford, OX3 7DQ, UK}
\address[kings]{Department of Biomedical Engineering, Division of Imaging Sciences and Biomedical Engineering, King's College London, St Thomas' Hospital, London,  SE1 7EH, UK}

\begin{abstract}
Rectal tumour segmentation in dynamic contrast-enhanced MRI (DCE-MRI) is a challenging task, and an automated and consistent method would be highly desirable to improve the modelling and prediction of patient outcomes from tissue contrast enhancement characteristics -- particularly in routine clinical practice. A framework is developed to automate DCE-MRI tumour segmentation, by introducing: \emph{perfusion-supervoxels} to over-segment and classify DCE-MRI volumes using the dynamic contrast enhancement characteristics; and the \emph{pieces-of-parts} graphical model, which adds global (anatomic) constraints that further refine the supervoxel components that comprise the tumour. The framework was evaluated on 23 DCE-MRI scans of patients with rectal adenocarcinomas, and achieved a voxelwise area-under the receiver operating characteristic curve (AUC) of 0.97 compared to expert delineations. Creating a binary tumour segmentation, 21 of the 23 cases were segmented correctly with a median Dice similarity coefficient (DSC) of 0.63, which is close to the inter-rater variability of this challenging task. A second study is also included to demonstrate the method's generalisability and achieved a DSC of 0.71. The framework achieves promising results for the underexplored area of rectal tumour segmentation in DCE-MRI, and the methods have potential to be applied to other DCE-MRI and supervoxel segmentation problems. 
\end{abstract}

\begin{keyword}
Parts-based graphical models \sep supervoxel \sep classification \sep segmentation \sep DCE-MRI \sep rectal tumour
\end{keyword}

\end{frontmatter}


\section{Introduction}

\noindent
Dynamic contrast-enhanced magnetic resonance imaging (DCE-MRI) shows promise for tumour characterisation and measuring response to therapy from tissue perfusion characteristics \citep{karahaliou2014ahl,gollub2012dce, Goh2007fic}.  In DCE-MRI, a series of volumes are acquired during and after the injection of a Gadolinium chelate based paramagnetic contrast agent (CA), such as Multihance\texttrademark, Omniscan\texttrademark, Prohance\texttrademark, or Dotarem\texttrademark. A bolus of intravenous CA circulates through the vascular system and perfuses into the extravascular-extracellular space (EES) of the tissue, resulting in tissue enhancement, which depends on the tissue type and vascularisation. 

\begin{figure}[ht]
\centering
\includegraphics[width=13cm]{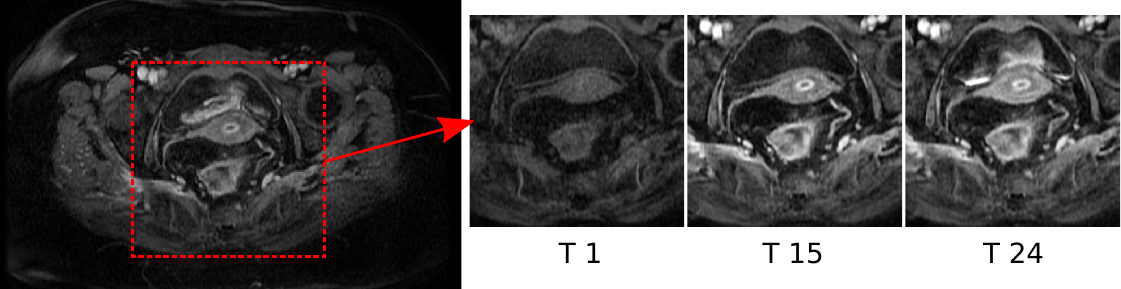}
\caption{A DCE-MRI axial slice at baseline, the 15\textsuperscript{th} and 24\textsuperscript{th} time point (Case 09).} \label{fig:exdce}
\end{figure}

\begin{figure}[ht]
\centering
\includegraphics[width=11cm]{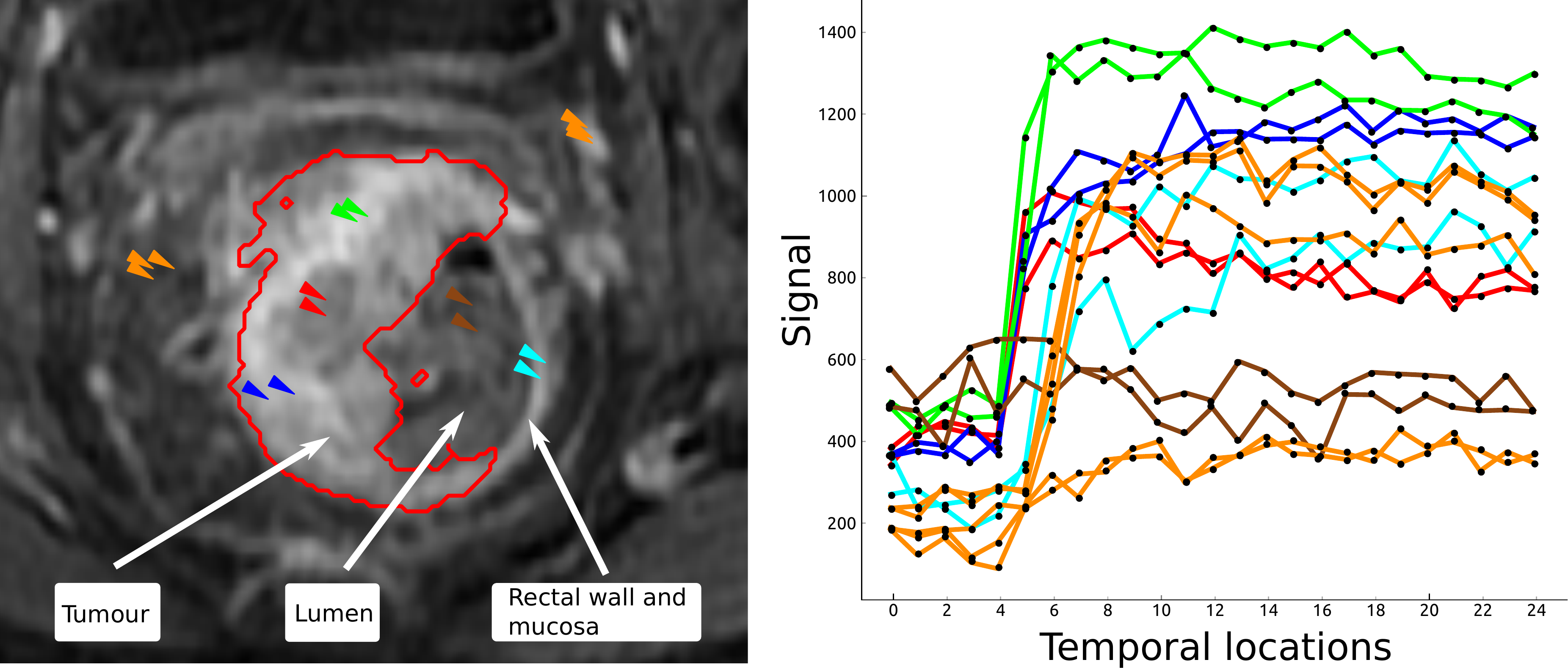}
\caption{Signal enhancement curves of the rectal tumour and surrounding tissue, which illustrates the heterogeneity in the tumour and surrounding region, as well as similarities in enhancement between local regions. \textit{Green}, \textit{red} and \textit{blue} are examples of enhancement for regions of the tumour, while \textit{cyan} shows the rectal wall and mucosa, \textit{orange} shows neighbouring tissue, and \textit{brown} the non-enhancing lumen.
} \label{fig:enhance}
\end{figure}

Neovascularisation, the formation of new vessels, is a key process of tumour growth, with angiogenesis being a common mechanism in colorectal tumours. This results in highly vascularised tumours with vessels that are thin, fragile and tortuous \citep{Goh2007fic}. The vasculature is also chaotic and leaky with regions of the tumour exhibiting hypoxia and necrosis, which leads to complex contrast enhancement patterns in DCE-MRI imaging of rectal tumours; Figure \ref{fig:exdce} shows an example of contrast enhancement in a rectal DCE-MRI scan.  Quantitative and semi-quantitative measures such as pharmacokinetic modelling can be used to analyse these perfusion characteristics in DCE-MRI \citep{Tofts1999ekp} and have shown the ability to assess the tumour \citep{Goh2007fic}. Heterogeneity in tumours can be further exploited to quantify tumour progression and response to therapy \citep{karahaliou2014ahl,gollub2012dce}. Figure \ref{fig:enhance} shows example enhancement curves for a rectal tumour and illustrates the heterogeneity of the tumour and surrounding region, while also demonstrating some consistency in tumour subregions.

To derive tumour-specific perfusion parameters, accurate delineation of the rectal tumour is required. It is difficult to visually distinguish tumour from surrounding structures on T1w sequences used for DCE-MRI; in clinical practice T2w imaging is preferred \citep{brown2005}, and can be registered to the DCE-MRI scan. DCE-MRI tumour delineation is further complicated due to the contrast-varying nature of the dynamic scan. However, even on T2w scans, manual segmentation of colorectal tumours is a difficult task, with substantial inter- and intra-rater variability \citep{franklin2014iiv,goh2008qac}, due to: partial volume effects in the axial plane; complex anatomy in the lower rectum, making it difficult to delineate normal anatomical structures; wall thickening due to venous congestion; and mucinous tumours. There is also often local motion between the T2w and DCE-MRI scans, which can lead to poor registration of the volumes, and therefore a poor delineation of the tumour in DCE-MRI.  An additional challenge in this study, and medical imaging in general, is the limited number of cases available. Therefore, a requirement of any automated DCE-MRI segmentation method is that it must be able to learn complex relationships efficiently from small datasets. This study is motivated by the rectal imaging trial (Churchill Hospital, Oxford, UK), which is an ongoing trial that aims to use pre-therapy DCE-MRI to assess patient outcomes after therapy. 

In this work, we propose an automated DCE-MRI segmentation approach that exploits features of the dynamic sequences to segment tumours, and provides a robust and consistent delineation for perfusion analysis. The key contributions of this method are: the introduction of \emph{perfusion-supervoxels} (supervoxels for dynamic contrast enhanced images, which are used to learn local characteristics of the tumour); and \emph{pieces-of-parts}, a novel formulation of parts-based graphical models (PGM) \citep{Felzenszwalb2005pso,Felzenszwalb2010} to directly improve supervoxel-based segmentations by including global anatomical constraints.

\begin{figure}[htp]
\centering
\includegraphics[width=12cm]{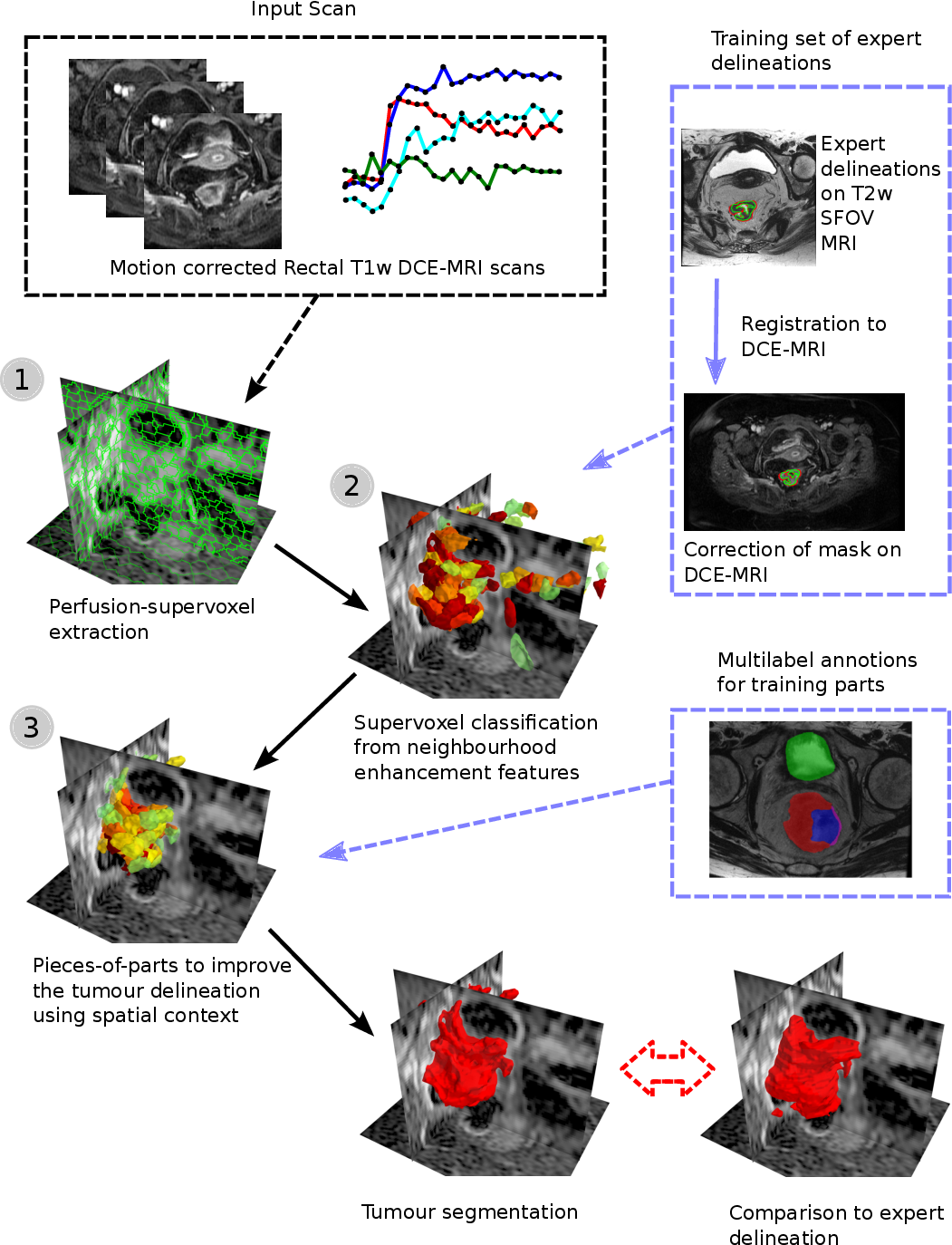}
\caption{Flowchart of the method for a new DCE-MRI volume. Steps 1 - 3 show the key steps of the framework.} \label{fig:pipeline}
\end{figure}

Figure \ref{fig:pipeline} shows a flowchart of the framework, where: a 4D contrast-enhanced DCE-MRI volume is acquired, the region is over-segmented using perfusion-supervoxels and each supervoxel is classified using learnt neighbourhood features. Pieces-of-parts then provides global constraints to the tumour segmentation using belief propagation on likely organ locations. Finally, the tumour segmentation is evaluated against an expert ground truth. This framework is applied to the novel and challenging task of automatic rectal tumour segmentation in DCE-MRI; the authors are not aware of any other method to automatically segment rectal tumours from DCE-MRI.

The initial steps in the development of this method were introduced in \cite{Irving2014act}, which includes the formulation of supervoxels for perfusion images and development of a supervoxel classifier. This is extended to include the novel pieces-of-parts graphical model. The validation is also improved using a more robust ground truth from expert delineations on multiple modalities and, while, previously, a number of cases required manual intervention, this method is evaluated in a fully automated fashion.

In the following section we place this work in the context of the state-of-the-art. Section \ref{sec:meth} outlines the method which is divided into perfusion-supervoxels (Section \ref{sec:superrep}), extraction of features (Section \ref{sec:featclas}) and pieces-of-parts (Section \ref{sec:parts}) which relate to Steps 1-3 in Figure \ref{fig:pipeline}. The dataset, experimental set-up and results are described in Section \ref{sec:eands}. Finally, the discussion is found in Section \ref{sec:discuss}.

\section{Background} \label{sec:prevmeth}

Pharmacokinetic models are a common method of extracting quantitative measures of perfusion from the contrast enhancement curves of DCE-MRI volumes \citep{Tofts1999ekp}. However, these measures depend on the selection of the compartment model, which in turn depends on the tissue characteristics, as well as the choice or measurement of an arterial input function \citep{Irving2013pea}. Tumour heterogeneity means that a single model is rarely appropriate for the whole tumour and surrounding region \citep{kallehauge2014tkm}. As an alternative to these model based methods, \cite{hamy2014rmc} apply principal component analysis (PCA) to the signal enhancement (SE) curves, and use this decomposition to distinguish motion from enhancement in a DCE-MRI based registration framework. In this study, we also take a data driven approach to avoid any assumptions about the underlying enhancement patterns in DCE-MRI; PCA is used to decompose the signal enhancement but is applied to supervoxel and feature extraction, unlike previously for registration \citep{hamy2014rmc}. 

DCE-MRI tumour segmentation, and in particular colorectal tumour segmentation, remains relatively unexplored. Breast DCE-MRI segmentation methods are more common, and the fuzzy c-means clustering approach is an established method \citep{chen2006fcm}. A benefit of this method is that it is unsupervised; the method iteratively assigns a fuzzy label to each voxel from the sum-of-squared distance of each enhancement curve to the cluster centres. \cite{McClymont2014} propose an alternative approach using mean shift clustering and graph cuts. Such methods have been effective for breast tumour segmentation; but the much more complex anatomy of the lower abdomen makes unsupervised clustering less effective compared to a supervised method, as demonstrated in our previous comparison \citep{Irving2014act}. 

Alternatively, key approaches in computer vision include the use of a superpixel representation with supervised learning and graph-based analysis. \cite{fulkerson2009cso} apply quick-shift clustering, and then objects of interest are detected using support vector machine classification of each superpixel, with conditional random fields (CRFs) to refine the segmentation. In medical imaging, \cite{mahapatra2013ads} also use supervoxel over-segmentation for detection of Crohn's disease in conventional 3D MRI, and \cite{Su2013} use superpixels with spectral clustering for brain tumour labelling. We also use a supervoxel and graph-based approach, but, unlike previous studies, we develop a method for 4D contrast imaging that uses: 1) a novel perfusion supervoxel over-segmentation and classification approach that exploits heterogeneous local neighbourhood characteristics; and 2) we developed an approach that couples supervoxels to a parts-based graphical model to efficiently include global anatomical relationships. Graph-based methods provide a convenient basis for incorporating these global anatomical relationships into our method. 

\subsection{Markov Random Fields and Parts-based graphical models} \label{sec:pgm}
Markov Random Fields (MRFs), and more specifically CRFs, are an important approach for refining image segmentations, where the image is represented as a graph of connected nodes, and the nodes are generally image pixels -- or superpixels (e.g. \cite{fulkerson2009cso}). The Hammersley-Clifford theorem guarantees (under certain assumptions) that the local interactions of an MRF can attain the global optimum, but it uses global constraints implicitly and weakly. A number of methods have attempted to explicitly include larger scale context information in the MRFs to capture both neighbourhood and global interactions. \cite{Shotton2009} incorporate texture-layout potentials into a CRF. This filter captures that layout of connected regions and is useful for scene labelling. However, this does not capture long range relationships between sparsely labelled regions that is required to exploit organ segmentations in large volumes. Alternatively, \cite{Torr2005} present a parts-based representation that includes global relationships between rigid parts in a moving object using part location, shape and texture information. Their method imposes strict constraints on the shape and relative location of each part, which is less suitable when individual parts are highly variable, as is the case for rectal tumours that vary both in shape and location relative to the colorectal anatomy. More recently, \cite{krahenbuhl2011eif} have introduced mean field inference on fully connected CRFs, which makes fully connected CRFs a feasible approach for capturing long range relationships in an image. This method shows improvements on previous methods but the formulation of the pairwise term using contrast sensitive two-kernel potentials cannot explicitly capture distance relationships between sparsely labelled regions e.g. organs. The efficient inference method is also not suitable for a superpixel/supervoxel representation because the feature space needs to be represented as a permutohedral lattice.

Alternatively, parts-based graphical models (PGMs), or pictorial graphical models, are an efficient method for landmark detection, and combine local features of individual parts with the spatial relationships between these parts \citep{fischler1973representation}. \cite{Felzenszwalb2005pso} introduce a Bayesian representation of the PGM by defining the graphical model as a graph $G=(\mathlarger{\mathlarger{\nu}}, \mathlarger{\mathlarger{\varepsilon}})$ where $\mathlarger{\mathlarger{\nu}}$ is the set nodes and $\mathlarger{\mathlarger{\varepsilon}}$ are the edges. The prior probability $P(L|I, \Theta)$ that the PGM has a certain configuration is given by:

\begin{equation}
P(L|I, \Theta) \propto \left( \prod_{i=1}^n p(I | l_i, u_i) \prod_{(v_i, v_j) \in \mathlarger{\varepsilon}} p(l_i, l_j | c_{ij}) \right) \label{eqn:pgm}
\end{equation}

where $p(I | l_i, u_i)$ is the probability of an image $I$ given that a part appearance parameters $u_i$ is at location $l_i$ and $p(l_i, l_j | c_{ij})$ is the likelihood that two parts are at locations $l_i$ and $l_j$ given the model parameters $c_{ij}$, and $\Theta=(u, c)$ are all model parameters. $l_i \in L$ are the locations of each part in the PGM. \cite{Potesil2014} have applied PGMs to medical images for landmark localisation of anatomy, and have shown that the spatial distributions can be learnt from small datasets making it suitable for medical images, where large datasets of images are often not easily available. 

A fully-connected CRF  could be incorporated into our framework. However, instead we introduce a pieces-of-parts approach -- a modified PGM adapted for the task of image segmentation. Instead of each supervoxel forming a node of a fully-connected CRF, the centre points of all supervoxels act as candidates (potential locations) for the nodes (tumour, lumen and bladder) of the PGM. The marginal distribution of the tumour is then calculated for these candidates using two-way belief propagation, and the parameters of the method are chosen in such a way that the marginal distribution provides an improved segmentation. This approach has a number of advantages (compared to fully connected CRFs) for segmentation problems with an expected spatial relationship between labelled regions of the image. First, only a small number training cases are required as distances between parts can be pre-learned on a few cases, and inference is straightforward using two-way belief propagation on the three nodes of the graph. Most importantly, pieces-of-parts can explicitly encode distance relationships and while a fully-connected CRF such as \cite{karahaliou2014ahl} only use distance to weight the importance of the connected nodes.

\section{Methods} \label{sec:meth}

The key steps of the framework are discussed in Sections \ref{sec:superrep}, \ref{sec:featclas} and \ref{sec:parts} and are shown as steps 1, 2 and 3 of Figure \ref{fig:pipeline}. Perfusion-supervoxels are introduced in Section \ref{sec:superrep}, which are used to over-segment the tumour and surrounding regions based on DCE-MRI perfusion characteristics. Features are extracted from each supervoxel and neighbourhood, and a classifier is trained to label supervoxels as tumour, lumen and bladder (Section \ref{sec:featclas}). This supervoxel segmentation approach provides a powerful tool for DCE-MRI tumour segmentation but uses only local analysis (supervoxel and nearest neighbours) and does not take advantage of relationships between various anatomical structures in the region. In Section \ref{sec:parts}, the supervoxels are reformulated as pieces-of-parts in order to propagate the belief about each supervoxel via a parts-based graphical model to update the tumour probabilities based on the surrounding anatomy.  

\subsection{Perfusion-supervoxel extraction} \label{sec:superrep}

Superpixel or supervoxel segmentation methods are an effective method of reducing an image into a set of locally similar regions, which reduces the complexity and redundancy of the image, and provides a more natural set of subregions for analysis. Previously, these methods have been used for greyscale or colour images, and we extend them to perfusion images in order to over-segment the volume from locally similar enhancement patterns.

\begin{figure}[t]
\centering
\subfloat[]{ 
\includegraphics[height=5cm]{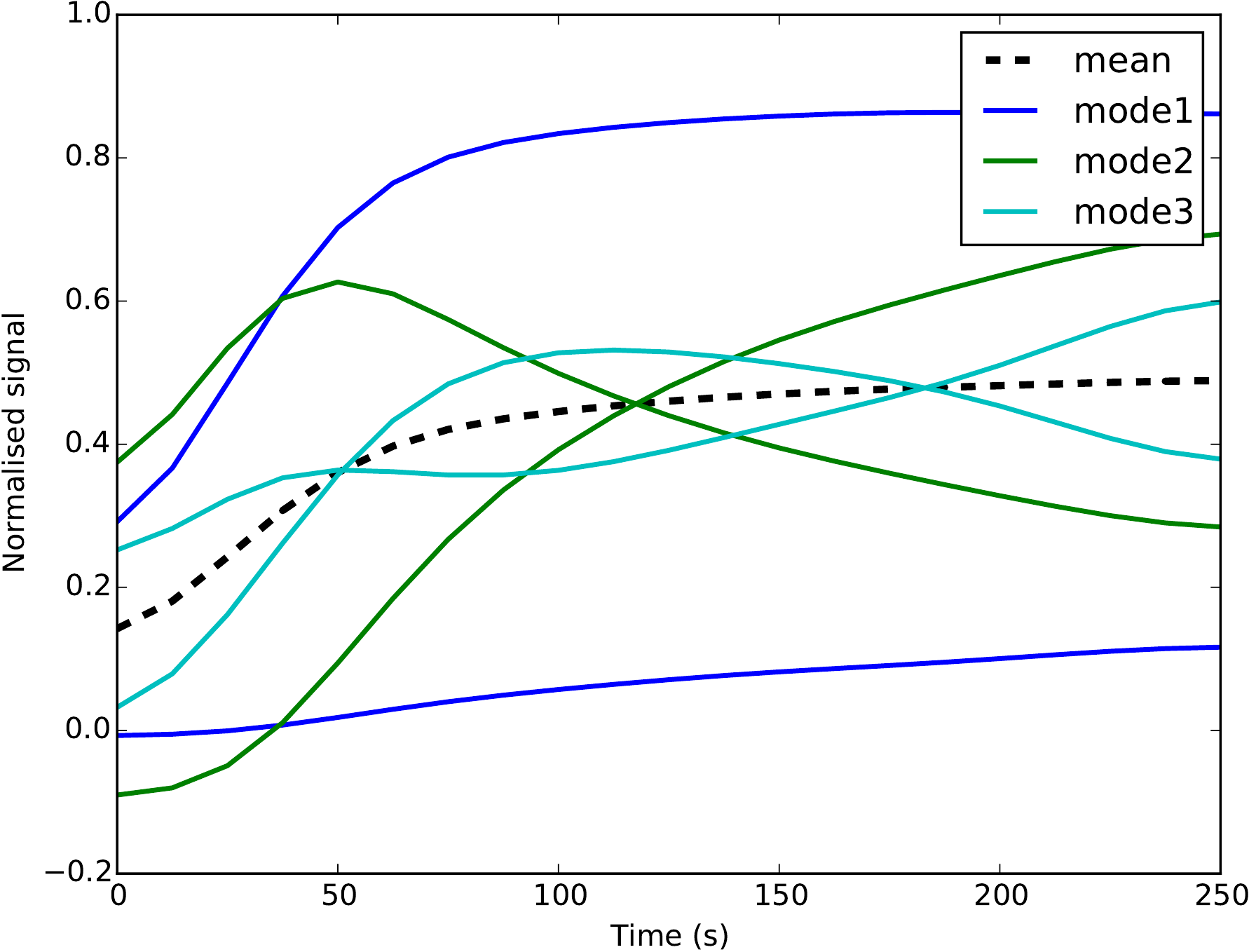}}
\subfloat[]{
\includegraphics[height=5cm]{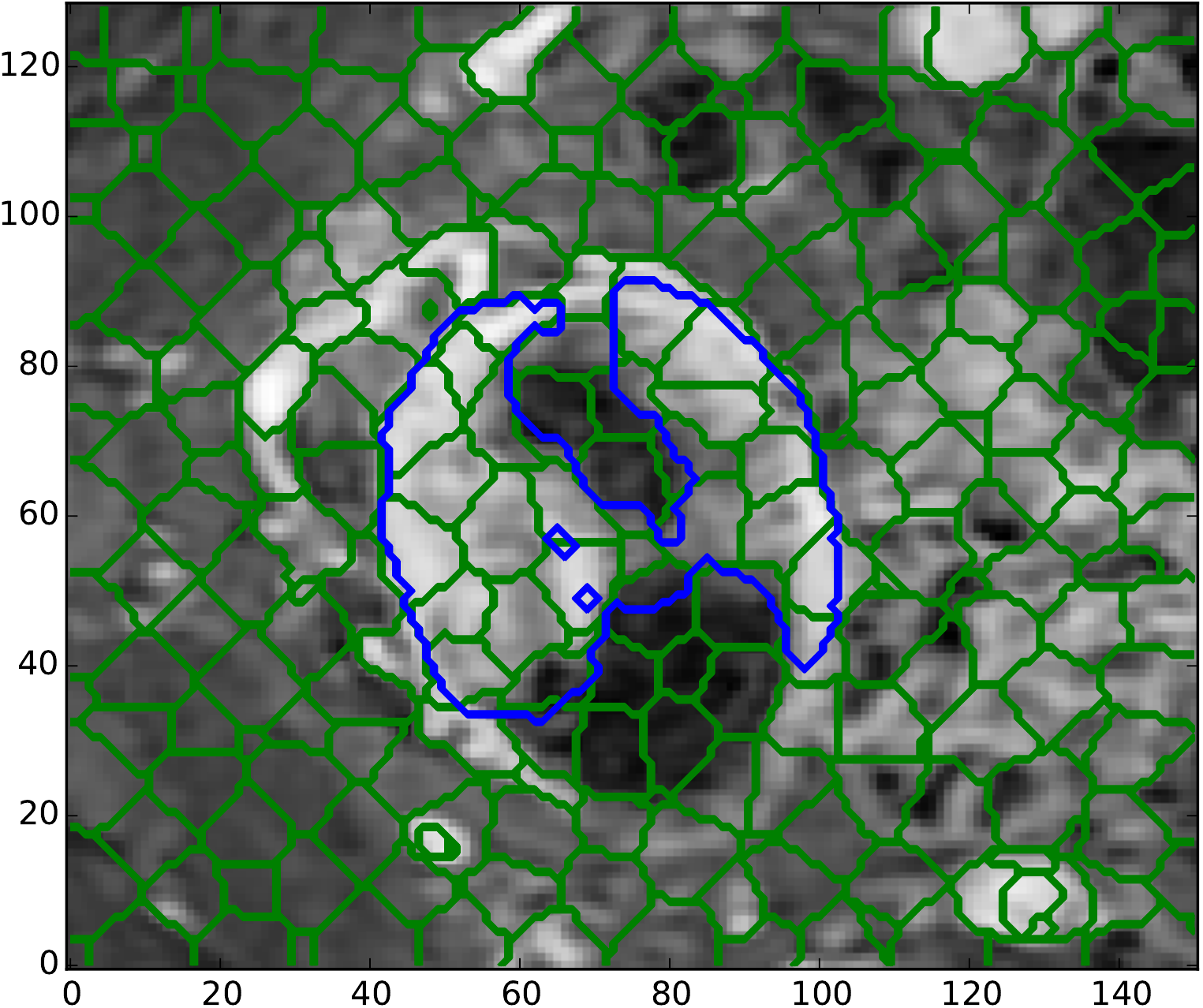}}
\caption{SLIC perfusion-supervoxel oversegmentation a) Variation of the first 3 modes of variation from the mean enhancement curve b) An axial cross section of the supervoxel oversegmentation shown  at a single time point of the 4D DCE-MRI volume ($S=350$ and $c= 0.11$). \emph{Blue} shows the ground truth segmentation .} \label{fig:slic}
\end{figure}

PCA is used to project a set of n-dimensional points into uncorrelated space using a linear transform and is commonly used for dimensionality reduction, and in our case is applied to the signal enhancement (SE) curves of the DCE-MRI. The transformed representation is given by $\mathbf{b}= \Phi^T (\mathbf{x}-\mathbf{\bar{x}})$, where $\mathbf{x}$ is a vector containing the SE curve values \emph{x(t)}, $\mathbf{b}$ is the representation in uncorrelated space and $\Phi^T$ is the transposed matrix of eigenvectors of the covariance matrix of the SE curves. The original SE curve can be reconstructed from the mean curve and the weighted sum of each principal component by $\mathbf{x} \approx \mathbf{\bar{x}}+\Phi \mathbf{b}$. The standard deviation of each mode is given by $\sigma_i = \sqrt{\lambda_i}$ where $\lambda_i$ are the eigenvalues of the covariance matrix. Figure \ref{fig:slic}(a) shows variation of the modes with one standard deviation from the mean for mode 1 ($\mu \pm \sigma_1$), and 3 standard deviations for modes 2 and 3 ($\mu \pm 3 \sigma_i$). The enhancement curves constructed using the first one, two, and five modes represents 97\%, 99\% and 99.9\% of the variance, respectively\footnote{Each curve is first smoothed with a 1D Gaussian filter of $\sigma = 1$}. 

Simple linear iterative clustering (SLIC) \citep{achanta2012slic} is a method for generating superpixels from colour or greyscale images using an adaptation of \emph{k}-means clustering that uses a distance function with both colour and distance similarity terms. The method is initialised by $k$ cluster centres that are sampled on a grid with separation $S = \sqrt{(N_v/k)}$, where $N_v$ is the number of pixels/voxels. Pixels are then assigned to each cluster centre based on the distance function and by searching a $2S \times 2S$ neighbourhood. We have chosen SLIC because of its speed and memory efficiency when dealing with large images \citep{achanta2012slic}. We extend SLIC to an arbitrary \emph{n}-feature image so that 3D supervoxel regions can be extracted from the contrast enhancement patterns of 4D DCE-MRI volumes. The distance metric ($D$) is defined in terms of a feature distance ($d_f$) and a spatial distance ($d_s$):

\begin{align}
d_f^2 = \frac{1}{n}\sum_k^n (b_{jk} - b_{ik})^2 \label{eqn:super1}\\
d_s^2 = \sum_k^3 (x_{jk} - x_{ik})^2 \label{eqn:super}
\end{align}

As discussed earlier, principal components are used to represent the complex enhancement patterns, where $b_{jk}$ is the \emph{k}\textsuperscript{th} principal component of the \emph{j}\textsuperscript{th} voxel, and $(x_{j1}, x_{j2}, x_{j3})$ denotes the location in 3D space of the \emph{j}\textsuperscript{th} voxel in mm. The distance metric ($D$) is given by:

\begin{align}
D = \sqrt{(d_f)^2 + \left(d_s/r\right)^2} \label{eqn:dist}
\end{align}

where $r = \frac{S}{c}$ represents the ratio of the average size / grid separation ($S$) and compactness ($c$) -- the influence of the spatial distance metric ($d_s$).  In this study, principal components were extracted for each voxel of the volume from the SE of all voxels within the volume, $n=3$ principal component features are used with the variation of the components are normalised between [0, 1]. 

As discussed, a standard implementation of SLIC \citep{achanta2012slic} was adapted to use PCA features with the modified distance function (Eq. \ref{eqn:dist}). As with the standard implementation, the method is initialised by placing cluster centres on a regular grid within the volume. Voxels are assigned the closest cluster centre using k-means, within a $2S \times 2S \times 2S$ distance of each cluster centre and with Eq. \ref{eqn:dist} as the distance metric. New cluster centres are computed and the process is repeated iteratively until the change in cluster centres falls below a tolerance. This method does not guarantee that the regions are contiguous, and in a post-processing step, small disconnected regions are reassigned to the closest cluster. Full implementation details of SLIC can be found in \cite{achanta2012slic}. Figure \ref{fig:slic}(b) shows a 2D axial slice of the 3D supervoxel over-segmentation of the tumour and surrounding region. In the following sections we call this method \emph{perfusion-supervoxels}. 

\begin{figure}[t]
\centering
\includegraphics[width=12cm]{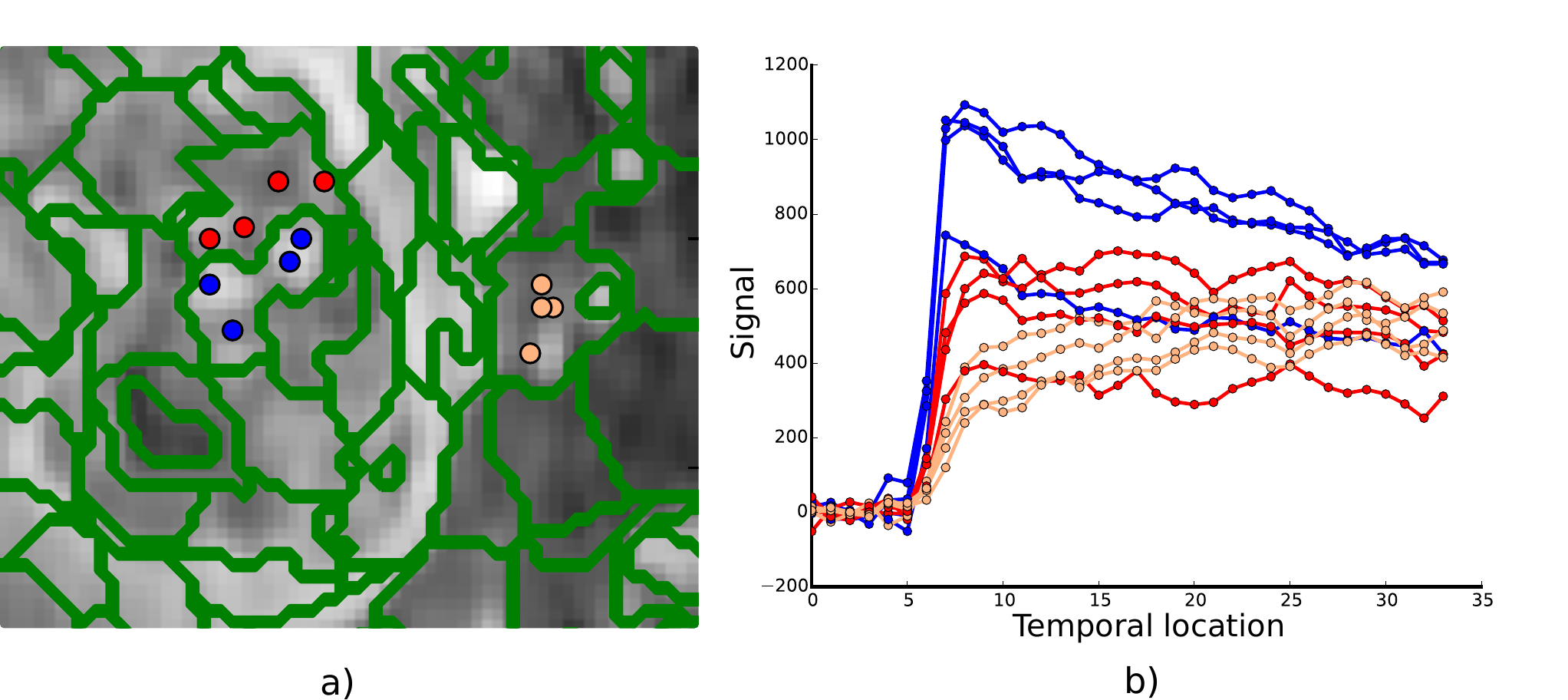}
\caption{Supervoxel representation with example enhancement curves for three supervoxel regions. Normalised PCA modes are used to extract supervoxels from enhancement curves to better represent curve shape. The magnitude of the curves can be thought of as being represented by a single PCA mode and, thus, the penalty is lower than if applied directly to the raw enhancement data. For example, a single blue curve may have the same magnitude as the curves of the red region but the shape is distinct.} \label{fig:pcacurv}
\end{figure}

As a side note, rather than using PCA, an alternative to the outlined approach would be to apply Eq. \ref{eqn:super1} and \ref{eqn:dist} directly to the SE curves. However, extending the supervoxel representation to dynamic scans using PCA provides a compact representation of each time series, extracts meaningful variation while reducing noise, and most importantly, when normalised, the supervoxel segmentation prioritises curve shape similarity over intensity similarity as shown in Figure \ref{fig:pcacurv}. If this normalisation were not performed, the supervoxel over-segmentation would be strongly influenced by the first mode (i.e. the maximum signal), which does not capture the complex enhancement patterns. 

\subsection{Supervoxel-based classification from neighbourhood enhancement features} \label{sec:featclas}
Having over-segmented the volume using perfusion-supervoxels, features are extracted from each supervoxel and the local neighbourhood. The tumour (as well as lumen and bladder) class probabilities are then assigned using a trained classifier.  

Principal components of the SE curves are used as features. First, a single patient scan is used to extract representative principal components and these are then used to project the enhancement curves of all other cases. The mean and standard deviation of the first five PCA components ($b_i$) were used to create a feature vector ($f_i$) for each supervoxel ($v_i$). Five PCA components are used during classification, instead of three used during clustering, because a trained classifier is more robust to poor features than an unsupervised clustering method. 

The supervoxel neighbourhood plays an important role in the detection of the tumour because it is likely to be surrounded by other heterogeneous tumour regions; lumen, that may contain non-enhancing air or stool; and thin rectal walls/mucosa (see Figure \ref{fig:enhance}). Therefore, an additional rotationally invariant set of features was developed to capture the enhancement of the supervoxel neighbourhood.  

\begin{figure}[t]
\centering
\includegraphics[width=6cm]{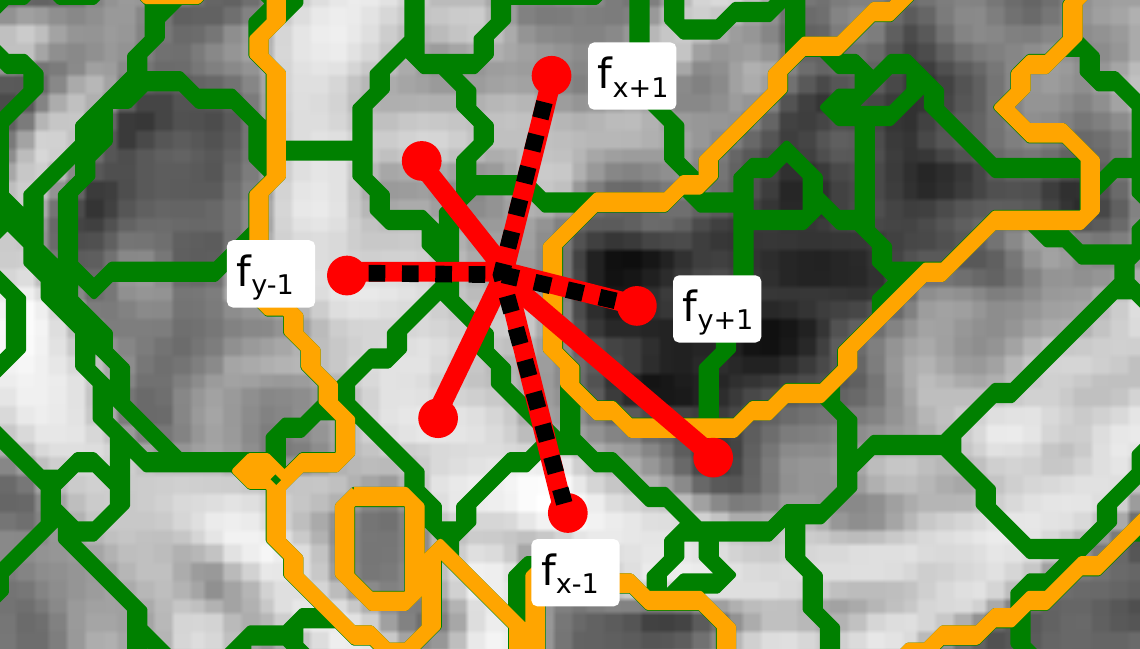}
\caption{Supervoxel neighbourhood (2D illustration). \emph{Orange} shows the tumour outline, with \emph{red} representing the neighbourhood connectivity for a single supervoxel. Dashed lines illustrate the neighbours that are most representative of the orientations from which an analogue to magnitude of the gradient can be calculated for each feature ($f_i$). } \label{fig:neigh}
\end{figure}

For this purpose, the supervoxel neighbourhood connectivity can be represented by an adjacency graph $G=(\mathlarger{\mathlarger{\nu}}, \mathlarger{\mathlarger{\varepsilon}})$  where $v_i \in \mathlarger{\mathlarger{\nu}}$ are supervoxels, and edges ($\mathlarger{\mathlarger{\varepsilon}}$) connect adjacent supervoxels $(v_i, v_j)$. A unit vector $\hat{d}_{ij}$ is constructed between the centroid of a supervoxel of interest $v_i$ and each neighbour $v_j$ to calculate the relative direction. A descriptor analogous to magnitude of the gradient is introduced to capture neighbourhood variation for each feature:
 
\begin{align}
f_{\bigtriangledown i}=\sqrt{(f_{i, x+1} - f_{i, x-1})^2 + (f_{i, y+1} - f_{i, y-1})^2 + (f_{i, z+1} - f_{i, z-1})^2}
\end{align}

where $f_i \in \mathbf{f}_i$ is the i\textsuperscript{th} feature from six neighbouring supervoxels that have a relative direction that is closest to that of the six orientations using $\hat{d}_{ij}$ and denoted by $[(x+1), (x-1), (y+1), (y-1), (z+1), (z-1)]$, as shown in Figure \ref{fig:neigh}.

The final supervoxel feature vector ($f_i$) is composed of 20 features: 10 (mean and variance of the 5 modes) from the supervoxel and 10 (magnitude of the gradient of the 10 features) from the local neighbourhood. Each feature in $f_i$ was normalised to [0, 1] over the entire training set and $f_i$ were used to train a linear discriminant analysis classifier to assign a probability of tumour, bladder and lumen to each supervoxel. We use the linear discriminant analysis to limit the amount of parameter selection required for small training sets. These class probabilities were used as the unaries for the parts-based model.

\subsection{Pieces-of-parts to improve tumour delineation using spatial context} \label{sec:parts}

The supervoxel classification method that is described in Sections \ref{sec:superrep} and \ref{sec:featclas} is used to detect and segment rectal tumours. These perfusion-supervoxels over-segment the tumour so that each tumour is accurately represented as a union of a number of supervoxels; and there are inevitably ``false positives" (supervoxel components that are labelled as part of the colorectal tumour but are at some distance from the colorectum).  The pieces-of-parts method addresses these issues by using the classifier class probabilities for each supervoxel in conjunction with the anatomical relationships in the scan to improve the segmentation. 

Previously, parts-based representations have been used for detecting points of interest or, alternatively, a bounding box surrounding an object of interest \citep{Felzenszwalb2005pso,Felzenszwalb2010,Potesil2014}. We extend this approach to a segmentation problem by exploiting the relationship between long range parts, but using this information to improve segmentation at a local supervoxel level, which we call a \emph{pieces-of-parts} segmentation method. The tumour, lumen and bladder are defined as parts (nodes of the graph) while supervoxels belonging to each part are candidate locations of the parts. This allows the global spatial relationships to be incorporated into the supervoxel segmentation.

\begin{figure}[ht]
\centering
\includegraphics[width=6cm]{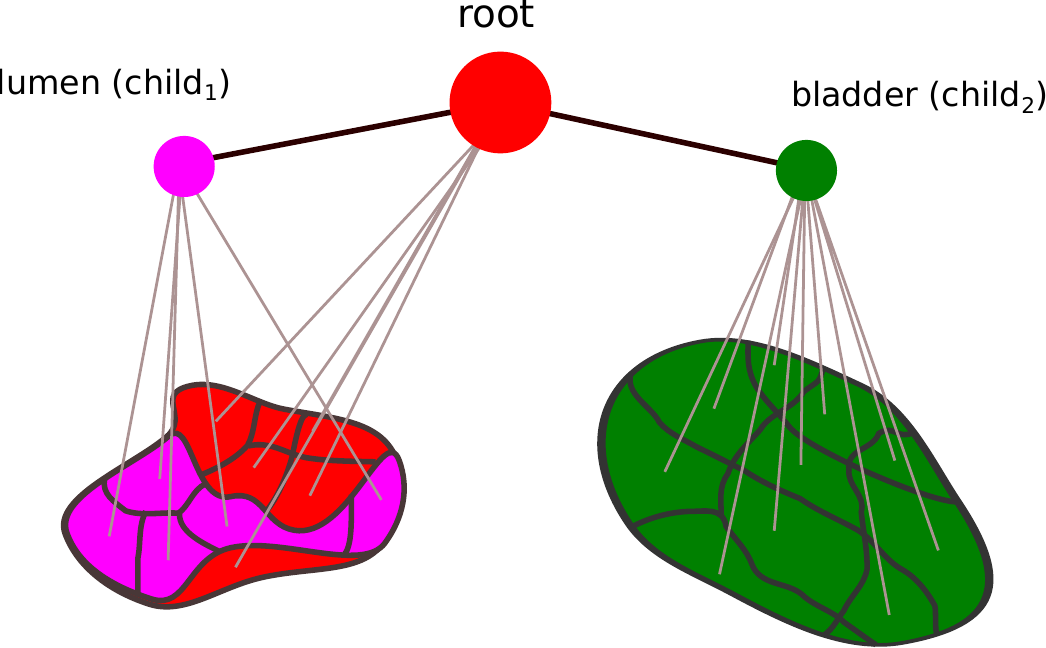}
\caption{Diagram illustrating the parts model where the tumour is defined as root and the lumen and bladder are children. Each part consists of pieces represented by supervoxels. } \label{fig:superpgm}
\end{figure}

Eq. \ref{eqn:pgm} can be reformulated into standard notation for a pairwise graph by changing $\psi_p(x_p) = p(I, l_i, u_i)$ and $\psi_{p, q}(x_p, x_q) = p(l_i, l_j | c_{ij})$ \citep{Murphy2012mlp}, which leads to:

\begin{equation}
p(x|v) = \frac{1}{Z(v)} \prod_{p \in \mathlarger{\nu}} \psi_p(x_p) \prod_{(p, q) \in \mathlarger{\varepsilon}} \psi_{p,q}(x_p, x_q) \label{eq:graph1}
\end{equation}

where $\psi_p(x_p)$ is the local evidence of node $p$ at position $x_p$, $\psi_{p, q}(x_p, x_q)$ is the pairwise potential between nodes at $x_p$ and $x_q$, and $Z(v)$ is the local normalisation constant. In the case of undirected edges the representation is the same for both PGMs and pairwise MRFs (or CRFS). However, for PGMs, the pairwise potential represents relationships between labelled parts of an image. In our case, the nodes represent the parts of our tumour model, including tumour, lumen and bladder as shown in Figure \ref{fig:superpgm}, and the supervoxels are candidates that may belong to one of the parts.

For inference, belief propagation (BP) is used for our \emph{pieces-of-parts} formulation (Section \ref{sec:pformulation}). BP is used to update the tumour (root) probabilities (Section \ref{sec:pcoll}), and lumen and bladder (child) probabilities (Section \ref{sec:pdist}) in the tree structure. Generation of the spatial prior distributions in this model are discussed in Section \ref{sec:pspatial}. The method makes no assumptions about the structure of each part involved other than the learnt spatial distributions, but in Section \ref{sec:pprobs} some problem-specific modifications are included. 

\subsubsection{Formulation of belief propagation for pieces-of-parts} \label{sec:pformulation} 

\begin{figure}[ht]
\centering
\includegraphics[height=3.5cm]{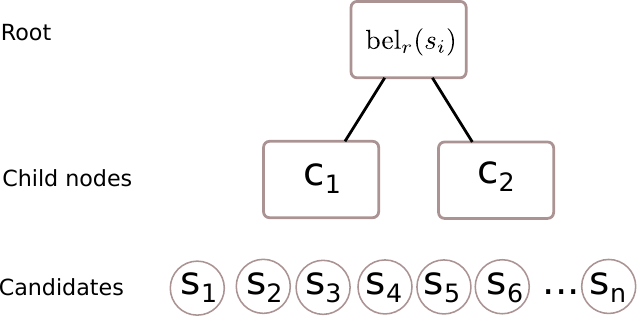} 
\caption{The hierarchy of the supervoxel pictorial graphical model. The tumour root ($r_1$) has child nodes ($c_1$, $c_2$) representing the lumen and bladder and these parts are informed by the beliefs of each candidate supervoxel in the volume ($s_1, s_2, ..., s_n$).} \label{fig:partdiag}
\end{figure}

The graph is a tree-structured PGM except that all supervoxels act as candidates for all parts. The first step of the BP collects the evidence from all candidate supervoxels that they belong to a part. This evidence from the candidates of each part is used to form a message that is propagated to the root, which is used to calculate the marginal distribution (or belief) $bel_r(s_r)$ for each root candidate $s_r$, with location $l(s_r)$ as shown in Figure \ref{fig:partdiag}. The evidence is then distributed to the child parts and used to update the hypothesis of each candidate belonging to each part. This is similar to the matching algorithm of \citep{Felzenszwalb2005pso} but instead of just determining the most likely candidate, we calculate the marginal distribution for all candidates. Therefore, the probability of each supervoxel ($s_r$) belonging to the tumour is determined by: 1) the unary potential of $s_r$ being tumour (based on the classifier from Section \ref{sec:featclas}), and 2) the beliefs about the likelihood of every other supervoxel in the volume belonging to a particular part, given $s_r$ is tumour.

\subsubsection{Collect evidence phase} \label{sec:pcoll}

In belief propagation, the belief of the root node, $bel_r(s_r)$, at supervoxel $s_r$, is given by the product of the local evidence, $\psi_r(s_r)$, and messages about the beliefs of the child nodes $m^-_{c \to r}(s_r)$ such that:

\begin{equation}
bel_r(s_r) =  \frac{1}{Z_r} \psi_r(s_r) \prod_{c \in ch(r)} m^-_{c \to r}(s_r)
\end{equation}

\begin{figure}[t]
\centering
\includegraphics[width=12cm]{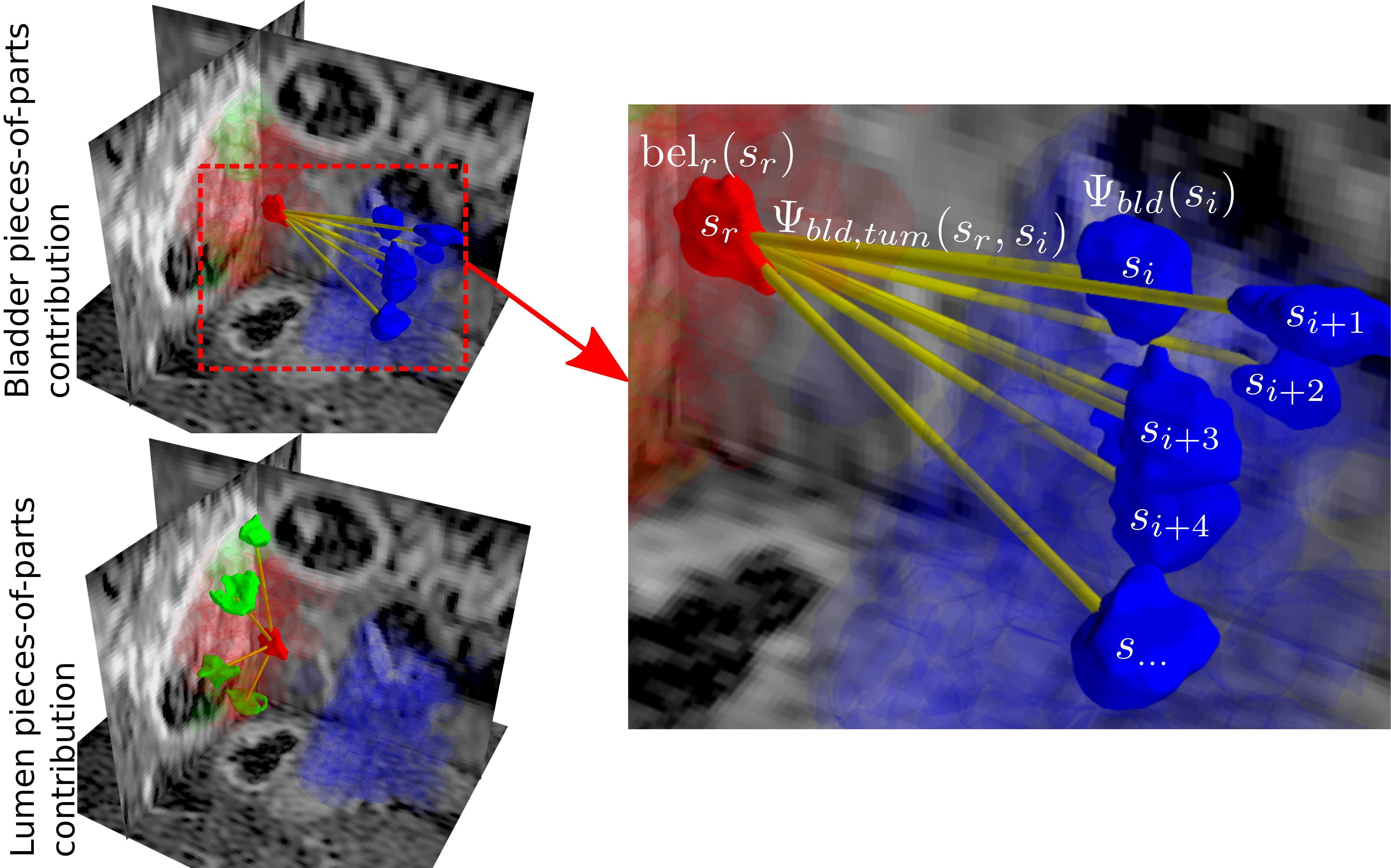}
\caption{Belief propagation on the supervoxel pieces-of-parts. Green, red and blue show regions belonging to the lumen, tumour and bladder, respectively. In the collect evidence phase, the likelihood of each other supervoxel belonging to either the lumen or bladder is used to update the belief of each potential tumour region. Yellow represents the spatial relationship between candidates that is used for belief propagation on the candidates and do not represent vertices of the graph.} \label{fig:collect_zoom}
\end{figure}

In our case, $ch(r) = \{ \mathrm{lumen}, \mathrm{bladder} \}$, and $s_r$ are not a sparse set of candidates but all supervoxels. The message $m^-_{c \to r}$ from each child candidate to the root is a weighted sum of the belief about each $s_c$ being a piece of a part given its position of $l(s_c)$, the root at position $l(s_r)$, and a potential of $\psi_c(s_c)$ of belonging to class $c$:

\begin{equation}
m^-_{c \to r} =  \sum _{s_c \in \hat{S}} \psi_{cr} (s_c, s_r) \psi_c(s_c)
\end{equation}

where $\hat{S} = \{\forall s \in S \land  c \neq r \}$ are all supervoxels, excluding the root candidate. $\psi_{cr} (s_c, s_r)$ is the probability that a root (tumour) candidate $s_r$ has distance $||l(s_r) - l(s_c)||$ from a bladder or lumen candidate (see Fig. \ref{fig:collect_zoom}).

\begin{figure}[t]
\centering
\includegraphics[width=9cm]{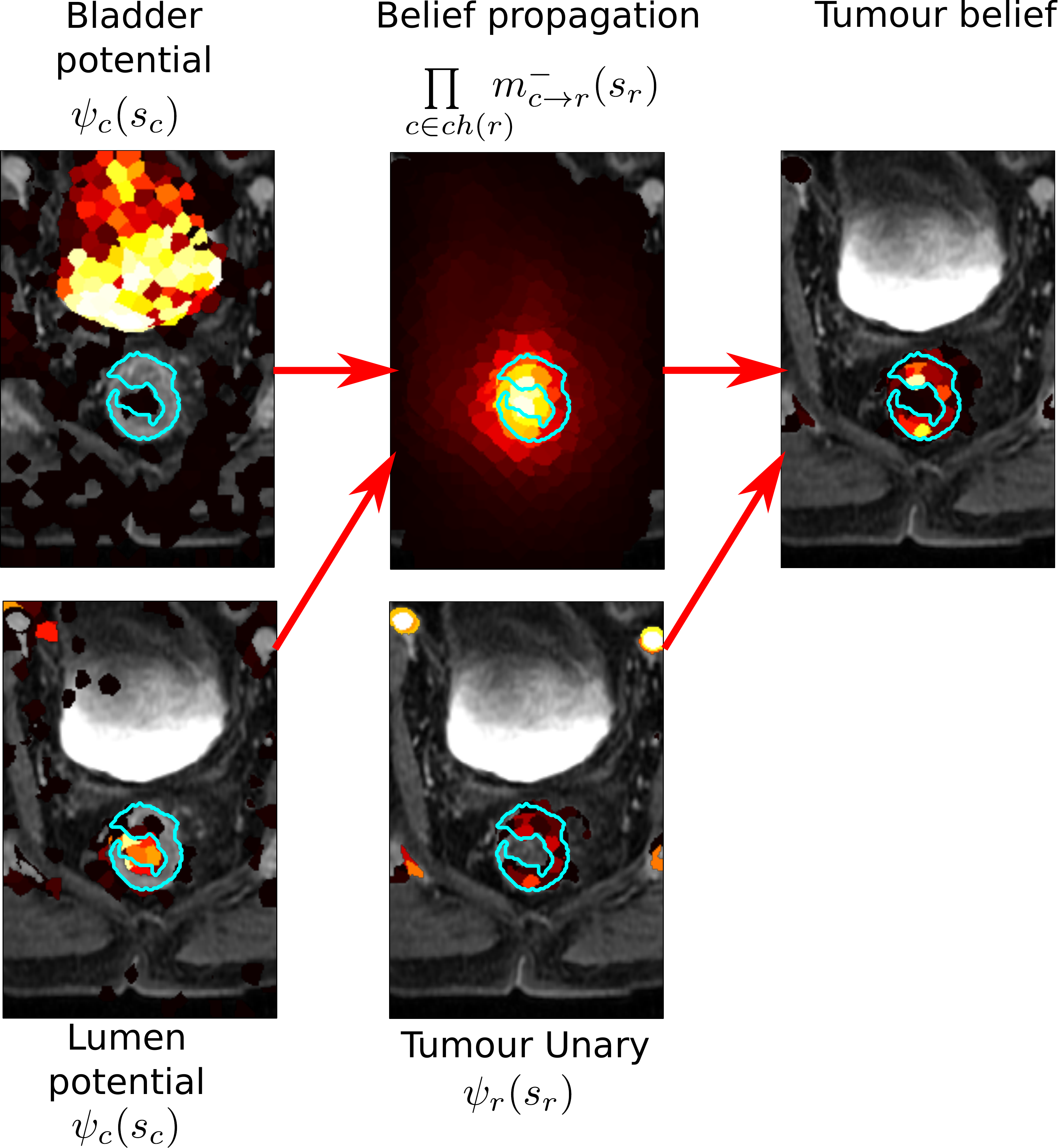}
\caption{Colormaps representing the potential of the bladder and lumen for each supervoxel (transparent is $<$ 0.01). Belief propagation is used to form $\prod_{c \in ch(r)} m^-_{c \to r}(s_r)$ for each supervoxel, which is combined with the tumour unary $\psi_r(s_r)$ to create a pieces-of-parts belief for the tumour location. This is compared to the ground truth (\emph{light blue}). Some bladder and lumen supervoxel probabilities might be incorrect but the majority vote leads to good tumour constraint.} \label{fig:partsteps}
\end{figure}

This step acts to collect evidence from all supervoxels based on the likelihood of being a component of child $c$ and their location relative to the root supervoxel candidate ($s_r$). $bel_r(s_r)$ can be interpreted as a belief that each supervoxel $s_r \in S$ of belonging to tumour, and can be used to create a tumour segmentation using an appropriate threshold. Figure \ref{fig:partsteps} shows the composition of the messages from the child parts $m^-_{c \to r}(s_r)$ and the local evidence $\psi_r(s_r)$ to form $bel_r(s_r)$. 

\subsubsection{Distribute evidence phase} \label{sec:pdist}
The belief about each child part can be distributed in a similar way using the beliefs we have about the root:	 $bel_c(s_c) = \psi_c(s_c) m^+_{r \to c}(s_r)$ where $m^+_{r \to c} = \sum_{s_r} \psi_{cr}(s_c, s_r) \frac{\mathrm{bel}_r}{m^-_{c \to r}(s_r)}$ is standard belief propagation \citep{Murphy2012mlp}.

\subsubsection{Spatial constraints} \label{sec:pspatial}

In this work we use a 3-dimensional normal distribution to represent the relative distance term between parts (lumen, tumour and bladder):

\begin{equation}
\psi_{cr} = \mathcal{N}(\mu_x, \mu_y, \mu_z, \sigma_x, \sigma_y, \sigma_z)
\end{equation}

\begin{figure}[ht]
\centering
\subfloat[]{
\includegraphics[width=5cm]{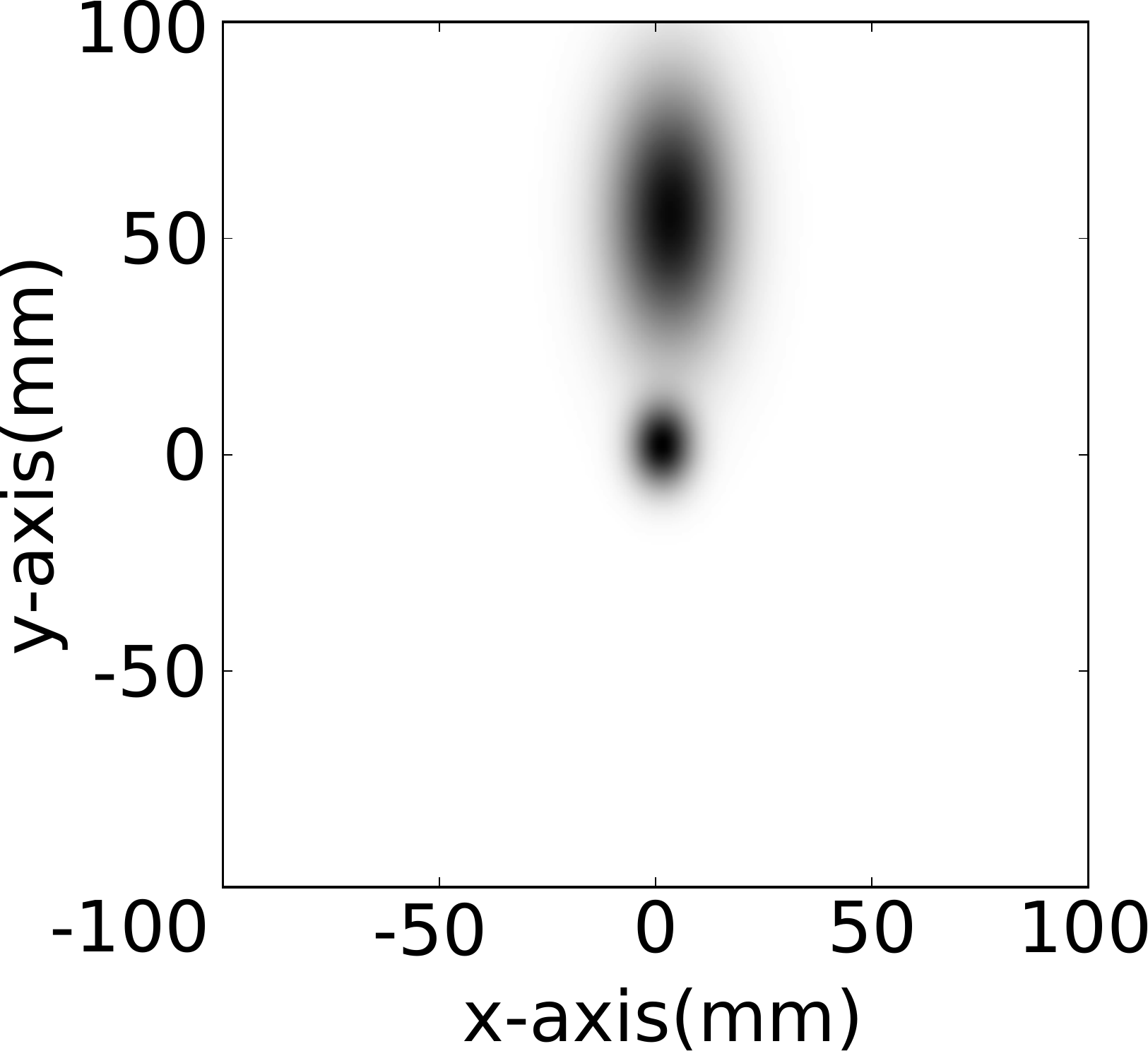}}
\subfloat[]{
\includegraphics[width=5cm]{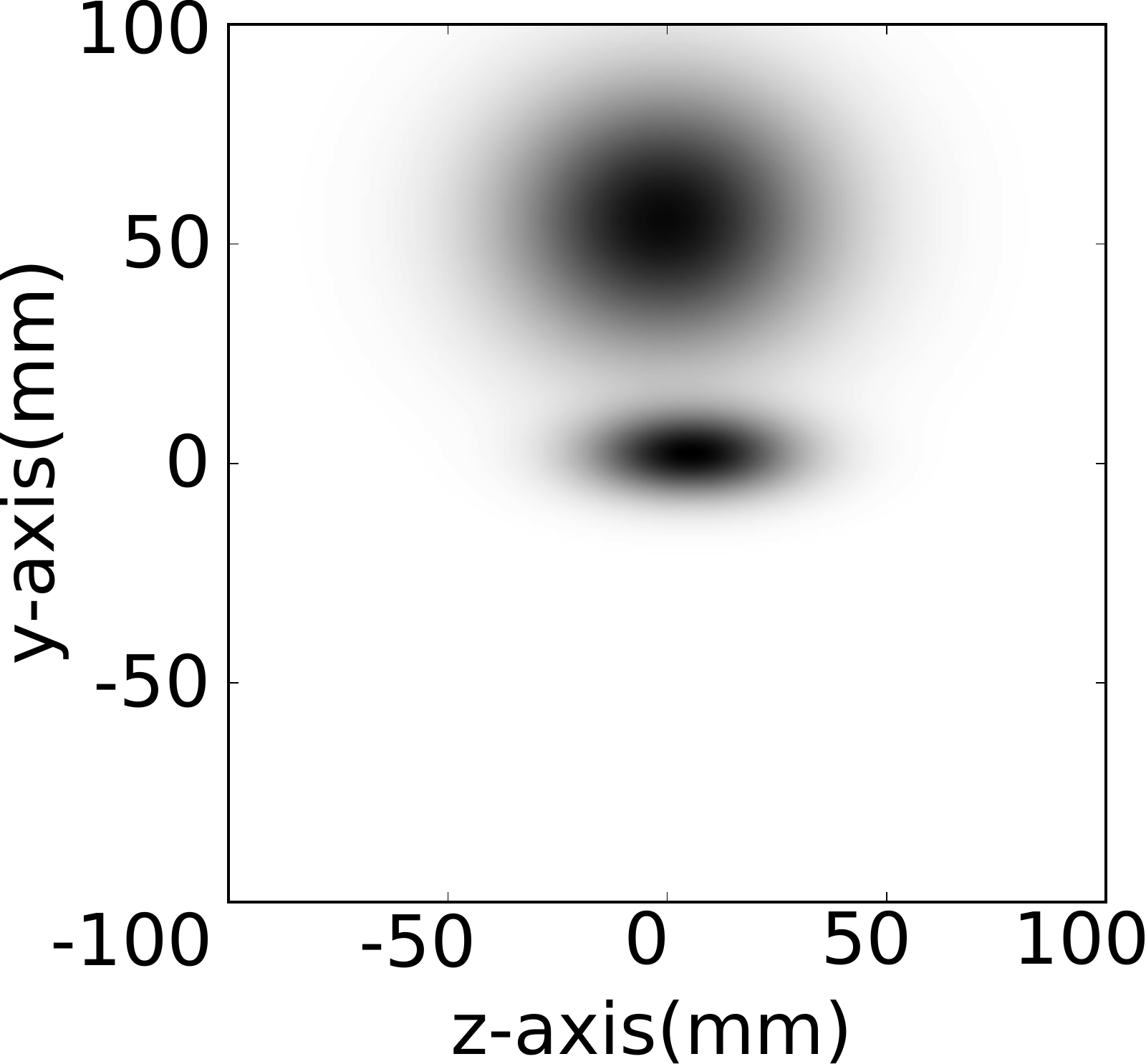}}
\caption{Normal distributions of the relative spatial locations of the child components: the bladder (top) and lumen (bottom) for the xy axis (a) and yz axis (b)} \label{fig:normdist}
\end{figure}

The parts are explicitly labelled in the training set and, therefore, we can train these spatial relationships independently of the classifier. While the size of each part influences the model, we find it sufficient to use the mean position of each part to calculate the distribution. Figure \ref{fig:normdist} shows the distribution and variance of the parts in 3D relative to the tumour centroid. Note that the lumen is in a similar location as the tumour but provides a strong constraint.

\subsubsection{Problem specific constraints} \label{sec:pprobs}

Our pieces-of-parts method can in principle be applied to any supervoxel segmentation problem that has a number of spatially related regions. However, in the case of DCE-MRI rectal tumour segmentation, we only keep the largest connected region after segmentation in order to exclude some smaller misclassifications. The bladder part was also constrained to the upper wall using a learnt part prior.   

\section{Experiments and Results} \label{sec:eands}

This section describes the application of our method to a rectal DCE-MRI dataset. The dataset is described in Section \ref{sec:mat}, with the experimental set up of the evaluation outlined in Section \ref{sec:ex} and Results presented in Section \ref{sec:res}. 

\subsection{Materials} \label{sec:mat}

Our method was evaluated on a rectal DCE-MRI dataset acquired as part of the Rectal Imaging Trial, Churchill Hospital, Oxford, UK between 2007-2014. T2-weighted small field-of-view anatomical MRI scans and T1-weighted DCE-MRI scans were acquired for 23 patients with stage 3 or stage 4 rectal adenocarcinomas. A 1.5T GE scanner was used with a gradient echo, fat-suppressed sequence (LAVA) (TR=4.5 ms, TE=2.2 ms and flip angle $12^o$). Patients then underwent downstaging chemo-radiotherapy. Multihance\texttrademark contrast agent was used, and images were acquired at approximately 12 \emph{sec.} intervals over 20 to 25 successive periods. The DCE-MRI voxel resolution was $0.78 \times 0.78 \times 2.0$ mm. T2w (resolution=$0.39\times0.39\times3.30$mm, TR=14ms, TE=12ms, flip angle $40^o$) are acquired in the axial-oblique plane, perpendicular to the long axis of the tumour to reduce partial volume effects. 

RHYTHM, a second smaller trial was also included to further demonstrate generalisability of the method. Scans were acquired in the same conditions as the first study at the Churchill Hospital, Oxford, and to date includes 4 cases with DCE-MRI scans.

Manual delineations of the tumour were used as the ``gold standard'' from which to train and evaluate our approach. DCE-MRI rectal tumour delineation is very challenging, even for the trained observer, and instead tumours were annotated on high resolution T2w anatomical scans, registered to the DCE-MRI, and further corrected to remove potential misregistration and rater error on the T2w, as follows:

\begin{enumerate}
\item Rectal tumours were delineated on T2w scans by a trained radiologist.
\item T2w scans were aligned and registered to the mid-volume of the DCE-MRI to correct for minor abdominal motion between scans using  \citep{heinrich2012min}.
\item The transformation was then applied to the delineations.
\item Delineations were further corrected on the DCE-MRI volume by the expert using the T2w MRI as a reference.
\end{enumerate}

\begin{figure}[t]
\centering
\includegraphics[width=10cm]{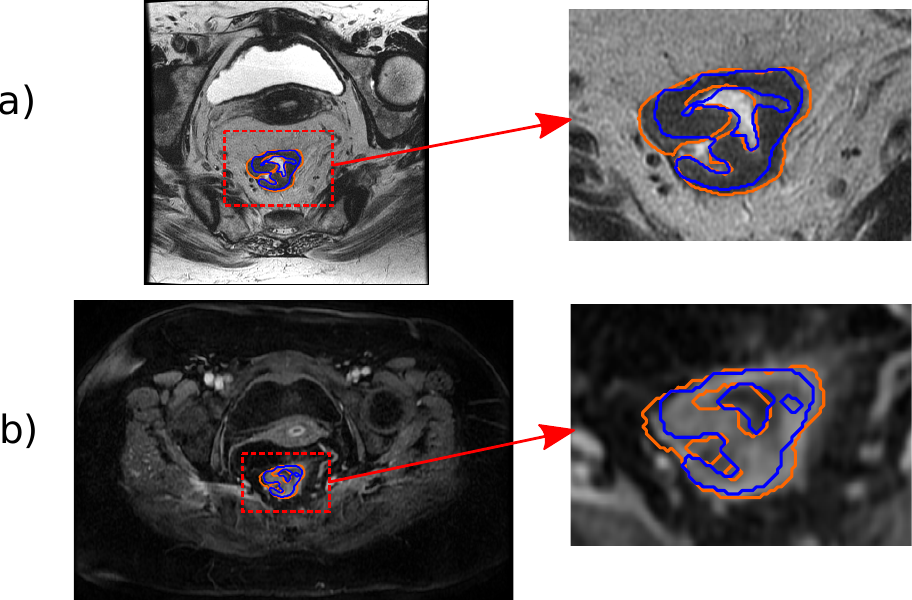}
\caption{Colorectal MRI scans (and cropped regions) with annotations to illustrate the inter-rater variability and registration method a) T2w axial-oblique MRI with tumour delineations from two experts (\emph{blue} and \emph{orange}) b) The corresponding DCE-MRI axial slice with registered annotations} \label{fig:scans}
\end{figure}

Fig. \ref{fig:scans} illustrates Steps 1-3, and also highlights the inter-rater variability, which sets a limit on the evaluation of our method against the radiologist ``gold standard". Inter-rater variability was calculated by asking two additional experts to annotate 10 cases, and comparing the results to the original annotator. The Dice similarity coefficient (DSC) were $0.73 \pm 0.13$ and $0.77 \pm 0.10$ \citep{Irving2014act,franklin2014iiv}, where DSC $\left( \frac{2|X \cap Y|}{|X|+|Y|}  \right)$ provides a measure of the region overlap between two sets of labels.  

\begin{figure}[ht]
\centering
\includegraphics[width=8cm]{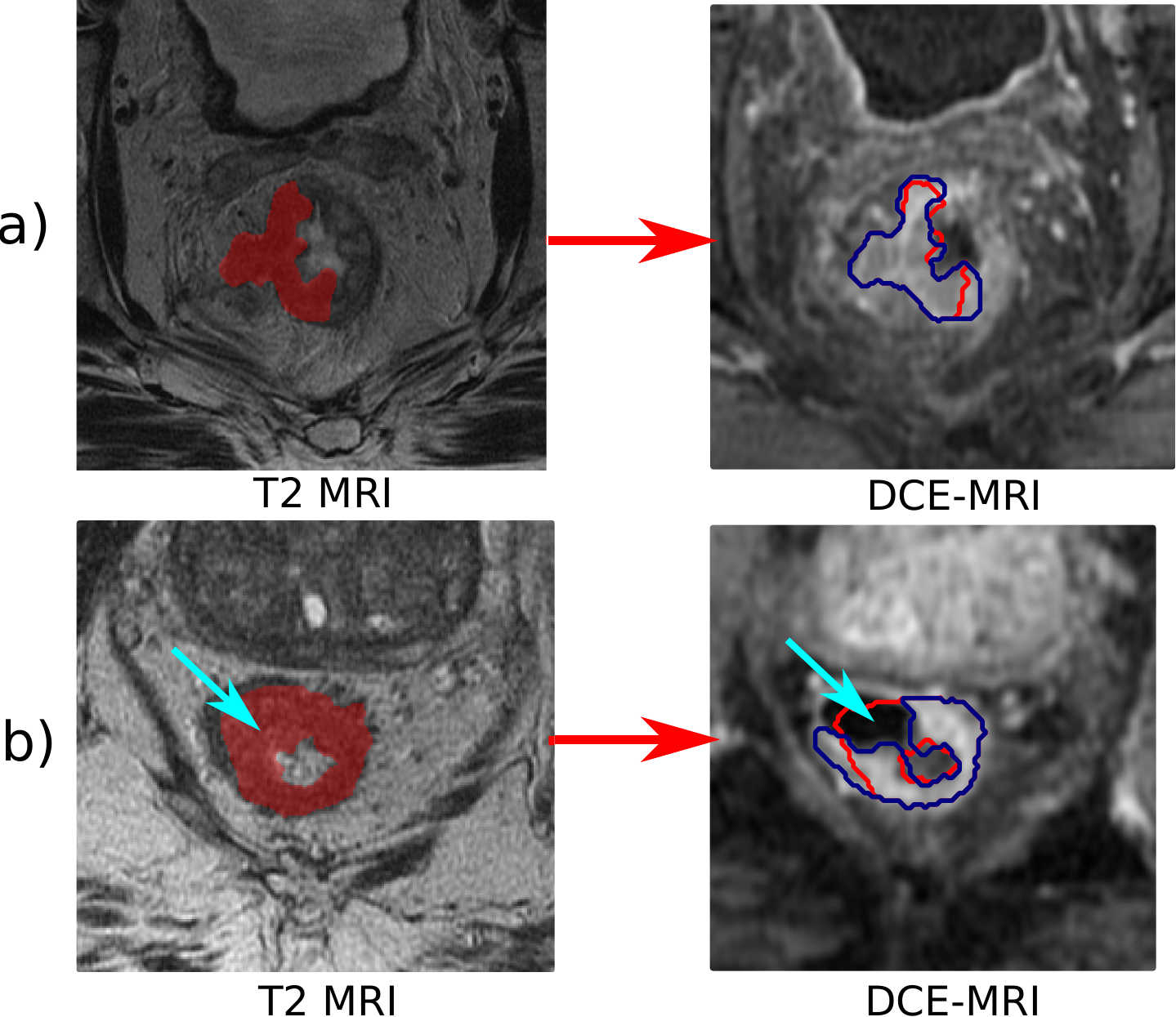}
\caption{Two examples (a and b) showing the T2w delineation (left), transformed mask (red) and the corrected mask (blue) on the DCE-MRI volume (right). The second example shows a very large local change to the tumour shape between scans, which requires relabelling of the registered volume. Blue arrows show the region that was originally mislabelled.} \label{fig:cor}
\end{figure}

\begin{figure}[ht]
\centering
\includegraphics[width=5cm]{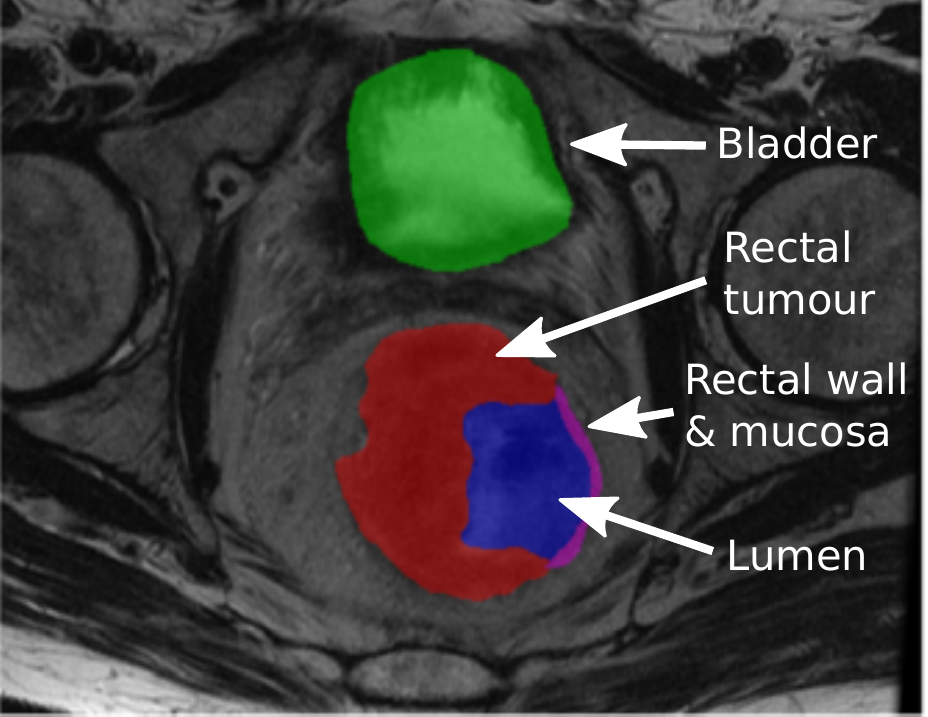}
\caption{Cross section and 3D rendering of the multilabel region of interest based on the T2w axial-oblique MRI. Green shows the bladder region, red the tumour, blue the lumen and magenta the lumen wall.} \label{fig:groundtruth}
\end{figure}

Fig. \ref{fig:cor} shows the manual correction in Step 4. Fig. \ref{fig:cor}(b) is an example where relabelling is particularly important because a region that had the appearance of tumour in the T2w scan, was found to be rectal wall from the DCE-MRI scan. The average DSC overlap between the registered and relabelled tumour volume for all cases was $0.91 \pm 0.06$ (with a minimum of 0.71 shown in Figure \ref{fig:cor}b). In addition to the expert tumour annotations, bladder and lumen labels were included for the pieces-of-parts training, as shown in Fig. \ref{fig:groundtruth}. This section further highlights the challenges of manual delineation and the need for a automated and consistent approach. 

\subsection{Experimental setup} \label{sec:ex}

\subsubsection{Preprocessing} \label{sec:pre}

\textbf{Region-of-interest}: Otsu thresholding is used to find a bounding box around the foreground of the MRI volume \citep{Otsu1979tsm}, and hard thresholds are applied as a simple preprocessing step to reduce the size of the region that is analysed by the framework. 1/3 of the width was removed from each side of the bounding boxing, 1/4 of the height from the top and 1/8 of the height from the bottom of the original region. Figure \ref{fig:exdce} has already shown an example of the reduced ROI.  

\textbf{Contrast-enhancement normalisation}: Contrast injection time relative to the start of the image acquisition has some variation between scans and the steepest gradient in contrast enhancement of the entire volume (contrast in the arteries) was used to detect injection time. SE was calculated by $SE = \left( \frac{\textrm{signal} - \textrm{baseline}}{\textrm{baseline}} \right)$, where the baseline is the mean intensity of each voxel for the time points before contrast injection. The curves were further normalised by the 80\textsuperscript{th} percentile of the maximum SE of the entire volume, which was designed to exclude the influence of blood vessels or late enhancement of the bladder in the normalisation. As there is also variation in the temporal resolution for some scans, linear interpolation was used to resample these scans to 12s intervals. 

\textbf{Motion correction}: We use an existing non-rigid image registration framework  based on the Diffeomorphic Demons modified with Normalized Gradient Fields (NGF) \citep{Haber2007,hodneland2014sdi}; details of the current implementation can be found in \citep{Papiez2014a}. The NGF is the contrast-invariant representation, which is used during estimation of the deformation vector field to handle intensity changes caused by contrast uptake, and has been previously shown to be suitable in DCE-MRI \citep{hodneland2014sdi}. The maximum number of iterations for each level is fixed to 25, and at each resolution level the standard deviation of the Gaussian smoothing was 2.8, 1.4 and 0.7 mm. These parameters were based on previous DCE-MRI datasets and are not changed for colorectal DCE-MRI. As expected, motion correction makes the segmentation more robust -- particularly at the edges of the tumour. However, the performance was poorer for some cases using just the perfusion-supervoxel classifier compared to previously \citep{Irving2014act}. This is because certain motion characteristics appear to have been learnt during classification. Therefore, global constraints, using pieces-of-parts, are even more important after motion correction because features from motion of the colon are excluded from the classification.

\textbf{Training mask pre-processing}: Expert delineations provide a voxelwise ground truth. We create a processed ground truth for training, which only has a single label assigned to each supervoxel. Supervoxels at segmentation borders might only partially overlap the expert segmentation (see Figure \ref{fig:slic}b). Therefore, to generate a single label, supervoxels with $\geq 50\%$ tumour labels are assigned as tumour. These are in turn used to reclassify supervoxels with an overlap $\geq 10\%$ as either tumour or background based on their similarity to other tumour regions in the volume using a linear classifier and mean PCA features, and provides a single label to each supervoxel for training.

\subsubsection{Post-processing} \label{sec:posproc}

The supervoxel and pieces-of-parts methods both assign a tumour probability to each supervoxel. These probability maps are thresholded to create a tumour segmentation. After creating this segmentation, only the largest connected region is included, and smaller disconnected regions are removed.

\subsubsection{Parameter selection} \label{sec:param}

\begin{figure}[t]
\subfloat[Case 01]{
\includegraphics[width=3.3cm]{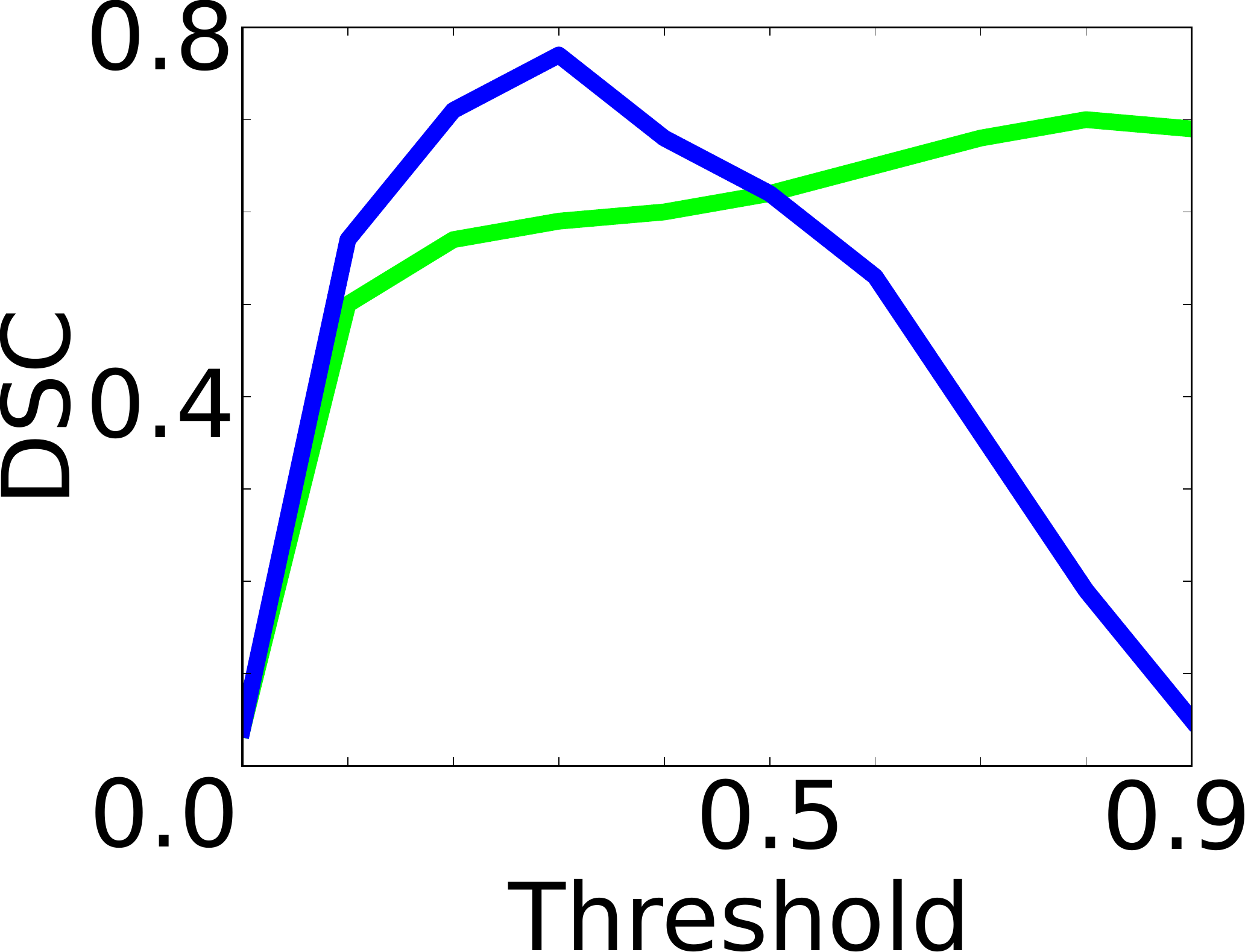}}
\subfloat[Case 03]{
\includegraphics[width=3.3cm]{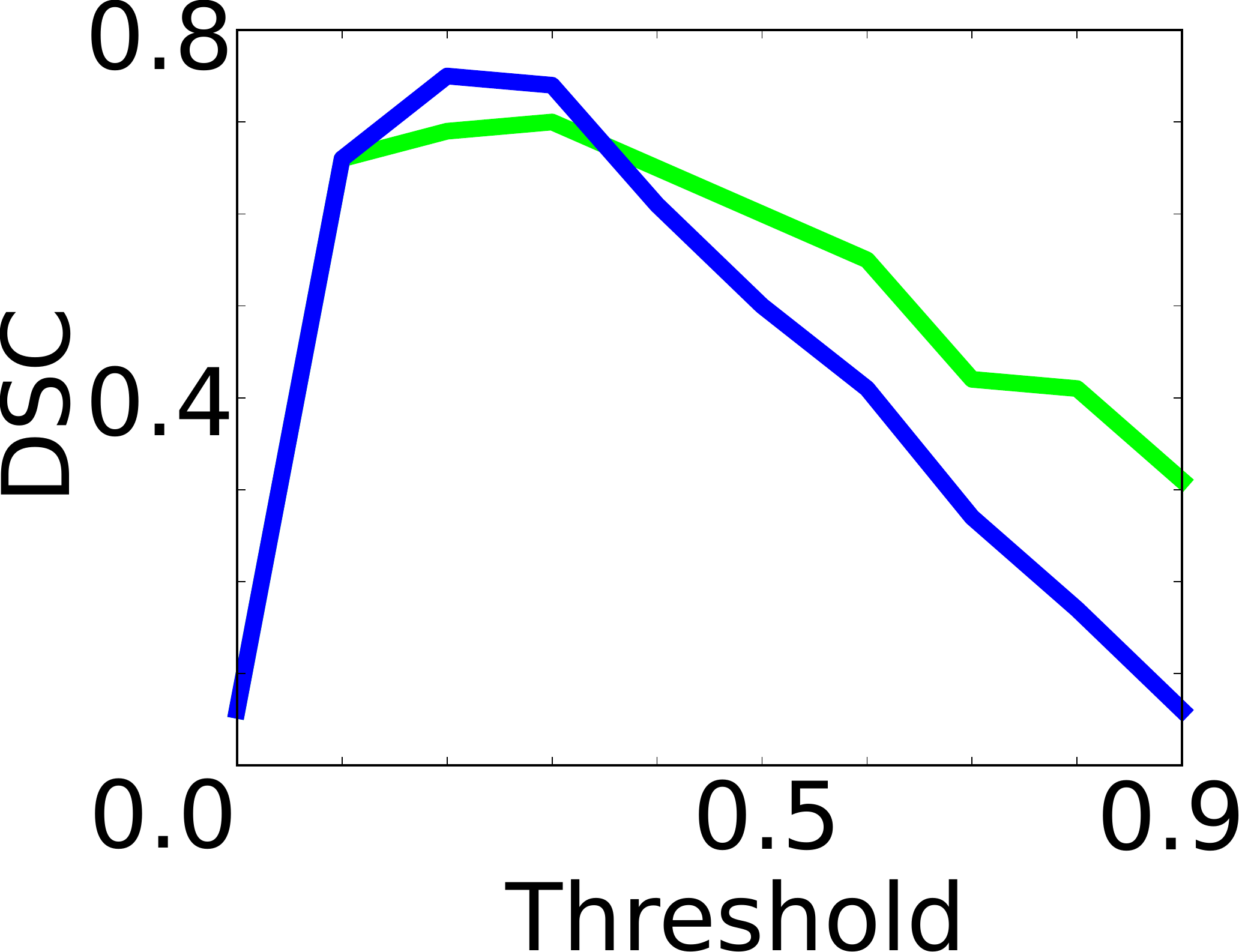}}
\subfloat[Case 04]{
\includegraphics[width=3.3cm]{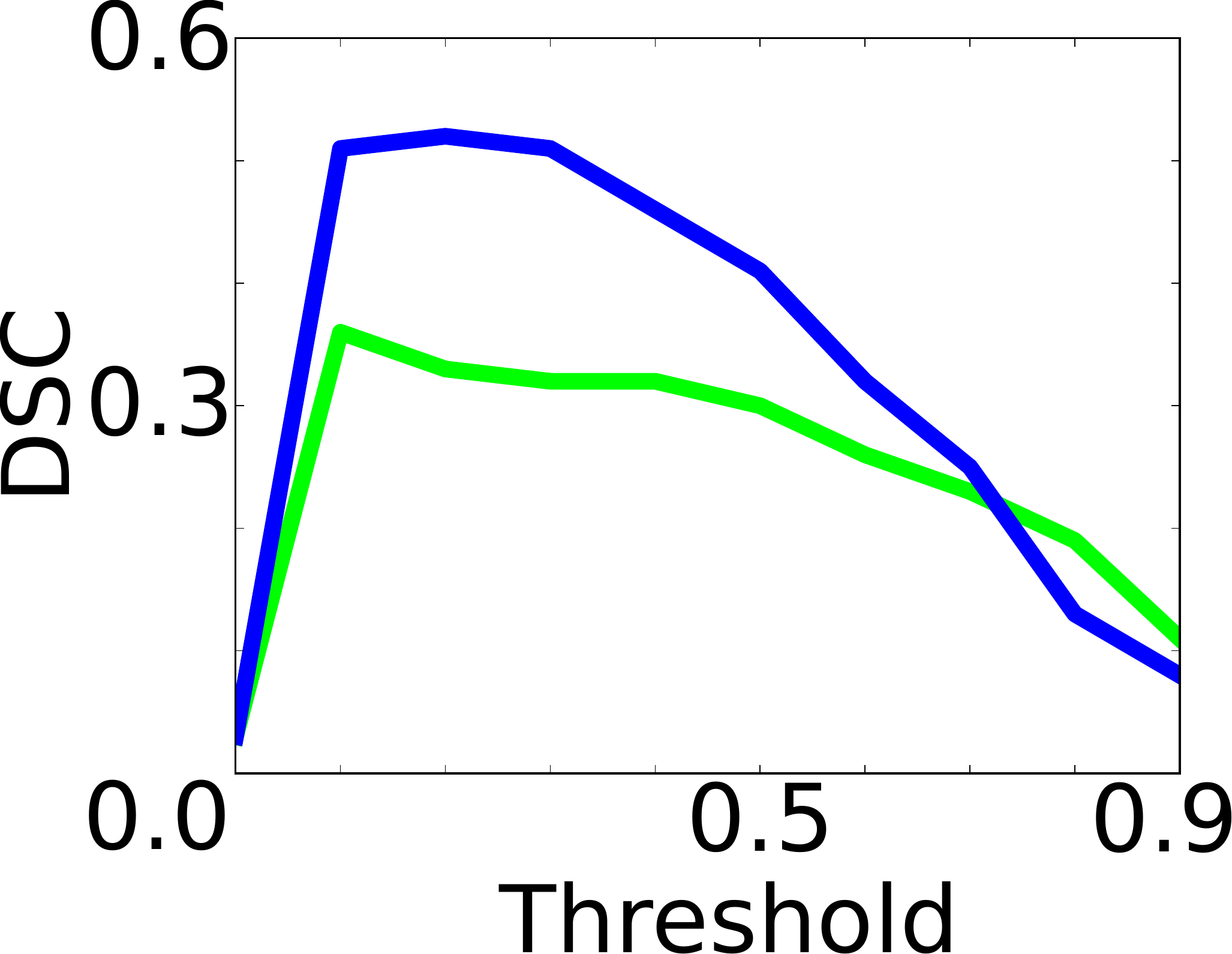}}
\subfloat[Case 06]{
\includegraphics[width=3.3cm]{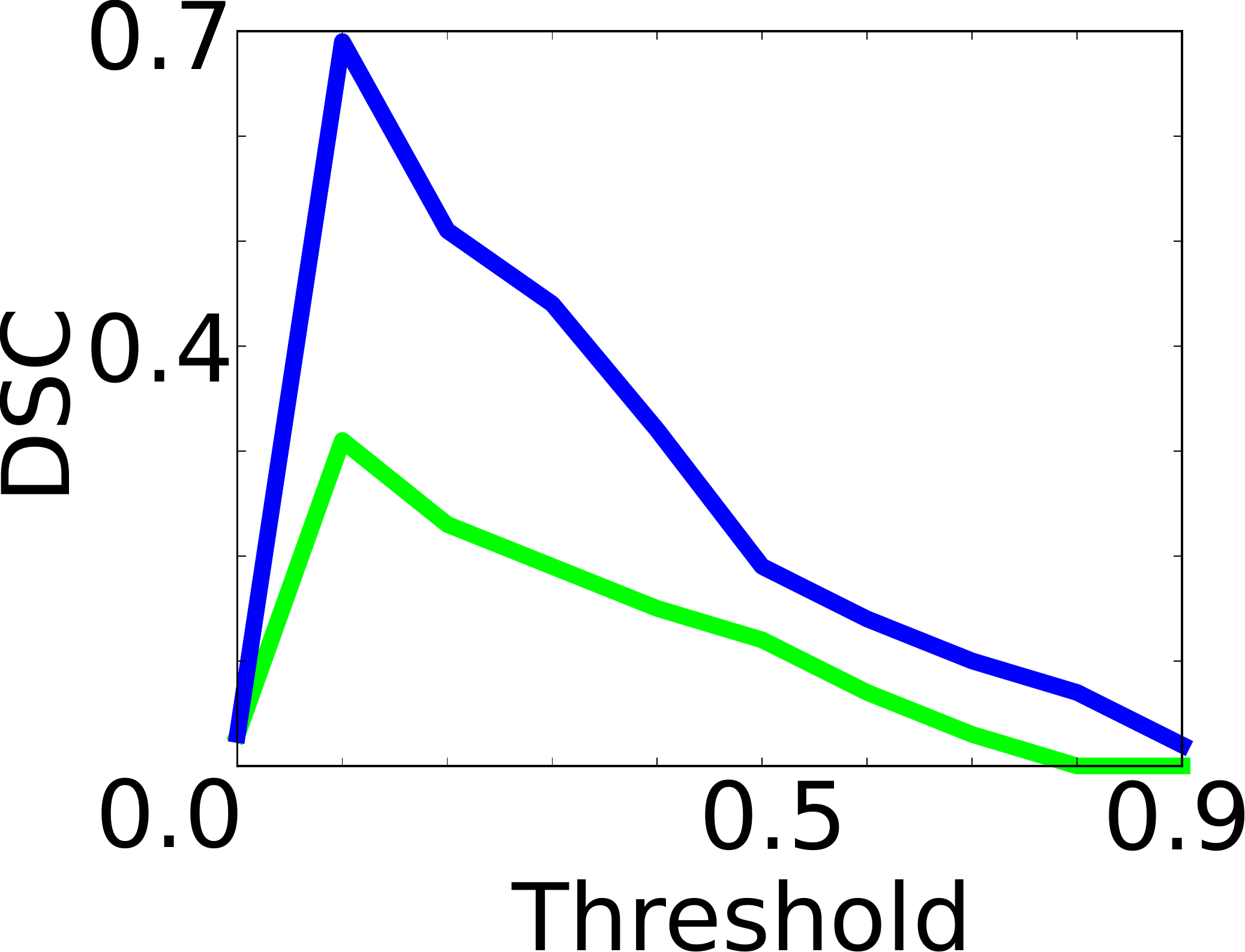}}
\caption{DSC as a function of the threshold for cases used for parameter selection (pieces-of-parts (blue) and supervoxels (green)). A threshold is chosen for both the supervoxels ($T_s$) and pieces-of-parts ($T_p$). } \label{fig:thresh}
\end{figure}

Given the limited data currently available for DCE-MRI, this method is designed to be effective and efficient for both large and small datasets. Therefore, we have attempted to limit the number of tunable parameters. The first four patients were used to find optimal parameters for the supervoxel method. The perfusion-supervoxel step requires two parameters (size $S_n$ and compactness $c$).  Note that the number of grid points ($n_{grid}$) and, therefore, $S$ is derived from the supervoxel size by $n_{grid} = N_v/S_n$. The linear discriminant analysis is as a parameter free classifier, where the output can also be given as a probability with optimal separating threshold ($T_s$) of 0.5. $T_s$ was verified on the four cases. Similarly, the \emph{pieces-of-parts} classifier only requires a single threshold parameter $T_p$ (equal weighting between the unary and pairwise terms was used). $T_p$ was also chosen from four cases, but includes a poorly performing case (Case 6), as shown in Figure \ref{fig:thresh}. Parameters are shown in Table \ref{tab:par}. These parameters are used for the key steps in the method. Other parameters, such as registration parameters, are discussed earlier. These parameters were also found to be fairly robust to the parameter choice. Supervoxel size ($S_n$), the mean number of voxels per supervoxel, in particular shows negligible change in accuracy in the range of 100 and 900 voxels.
\begin{table}[t]
\centering
\footnotesize
\begin{tabular}{lc}
\hline
 perfusion-supervoxels: & $c=0.05$ \\
 & $S_n=350$ \\
 & $T_s = 0.5$\\
pieces-of-parts: & $T_p =0.15$\\
\hline
\hline
\end{tabular} 
\caption{Algorithm parameters} \label{tab:par}
\end{table}

\subsection{Results} \label{sec:res}
The method was evaluated using leave-one-patient-out cross-validation on the DCE-MRI rectal tumour dataset. The results are presented from both a detection perspective using a receiver operating characteristic (ROC) and a segmentation perspective using the Dice similarity coefficient (DSC). 

\begin{figure}[h!]
\centering
\includegraphics[width=14cm]{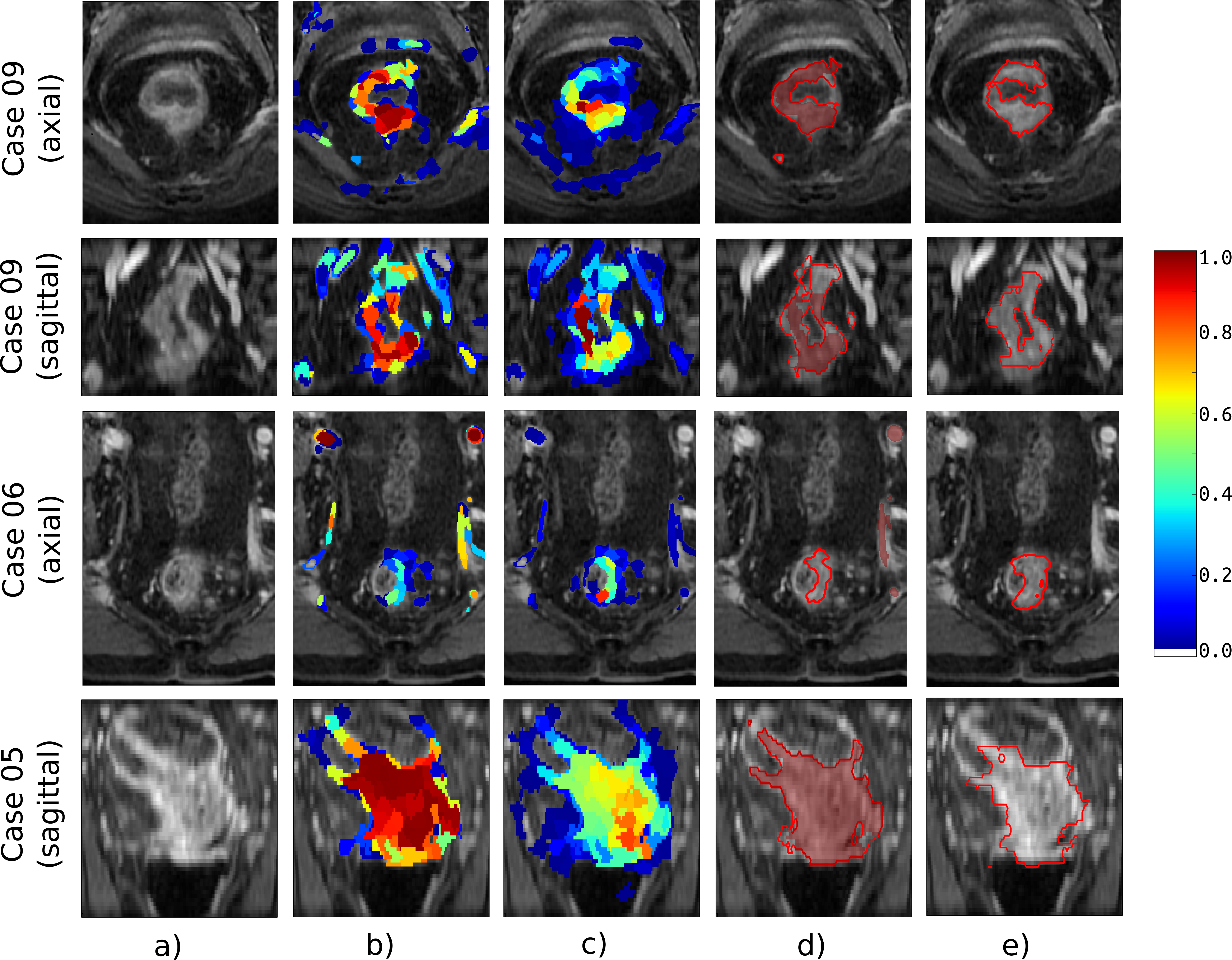}
\caption{2D cross sections of the 3D automatic segmentations of rectal DCE-MRI cases. Each row shows outputs of the framework and the final segmentation. (a) Unlabelled axial and sagittal cross sections through the DCE-MRI volume after contrast enhancement; (b) using perfusion-supervoxels, each supervoxel of an unseen case is assigned a probability of being tumour (shown by the color overlay where less than 0.01 is transparent) and then (c) improved using the pieces-of-parts graphical model. This is used to generate a segmentation by thresholding the probability maps (using thresholds $T_s$ and $T_p$ from Sec. \ref{sec:param}) and removing small disconnected regions. The final segmentations are shown in d) where the shaded region is derived from the perfusion-supervoxels probability in (b) and the outline is derived from the final pieces-of-parts probabilitity in (c), and compared to (e), which is the ground truth expert annotation.} \label{fig:examples}
\end{figure}

Examples of the classifier output and final segmentations are shown in Figure \ref{fig:examples}. Case 09 is an example of when the \emph{pieces-of-parts} method considerably improves the \emph{perfusion-supervoxel} segmentation while Case 05 is an example of an already good segmentation that is barely improved. Case 06 shows an example of a case which is misclassified by the original \emph{perfusion-supervoxel} algorithm but including \emph{pieces-of-parts} leads to a good segmentation because of the spatial priors; at temporal resolutions of 12 \emph{sec.}, large vessels may have similar characteristics to enhancing tumours, leading to the original misclassification. The DSC of this case is roughly equivalent to the median DSC and so is a representative example of the capabilities of the algorithm. 

\subsubsection{Experiment 1: Tumour detection}
\begin{figure}[h!]
\centering
\includegraphics[height=6cm]{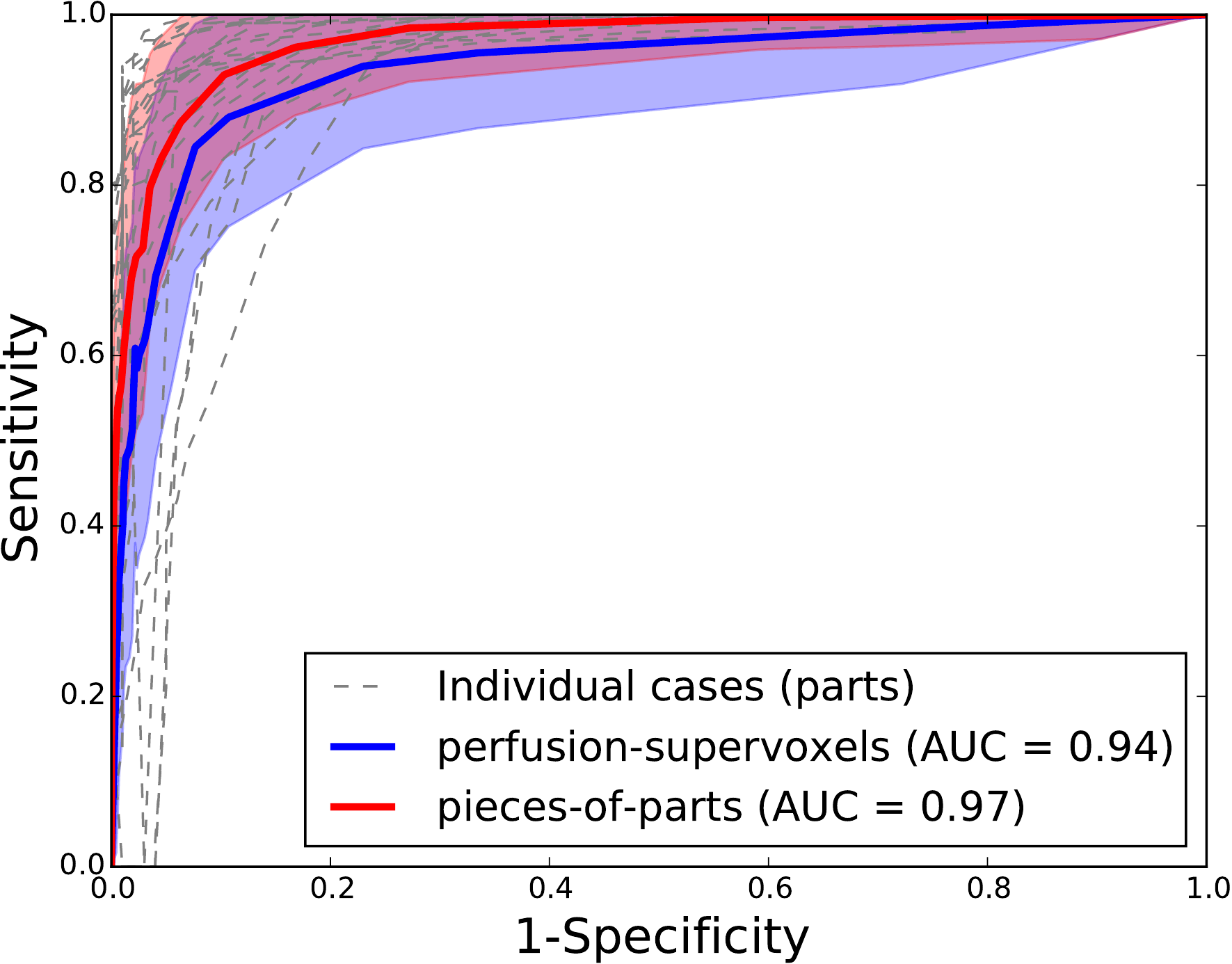}
\caption{Mean ROC curves with 95\% confidence intervals for the perfusion-supervoxel classification (blue) and the entire pieces-of-parts method (red). Taking the mean for all cases at each threshold. Pieces-of-parts ROC curves are also shown for the individual cases (gray). } \label{fig:roc}
\end{figure}

The classifier returns a probability for each supervoxel, and treating this problem as a voxelwise detection problem, a ROC curve can be generated.  Every voxel is labelled as tumour or background, and compared to the ground truth using sensitivity ($ \frac{TP}{TP + FN} $) and specificity ($ \frac{TN}{TN+FP} $), generated over a range of thresholds to obtain ROCs for each case and the mean ROC with 95\% confidence intervals. As can be seen in Figure \ref{fig:roc}, this method achieves a high accuracy (AUC = 0.97), and improved on the first step of this method (AUC = 0.94). These scores show that the tumour can be accurately detected with minimal false positive regions. 

\subsubsection{Experiment 2: Tumour segmentation}
Considering the results from the segmentation perspective, we want to provide an alternative segmentation approach to manual delineation of the rectal tumour by experts. DSC was used to evaluate the accuracy of the segmentation. Unlike the AUC score, DSC does not account for the background being correctly labelled. 

\begin{figure}[h!]
\centering
\includegraphics[width=12cm]{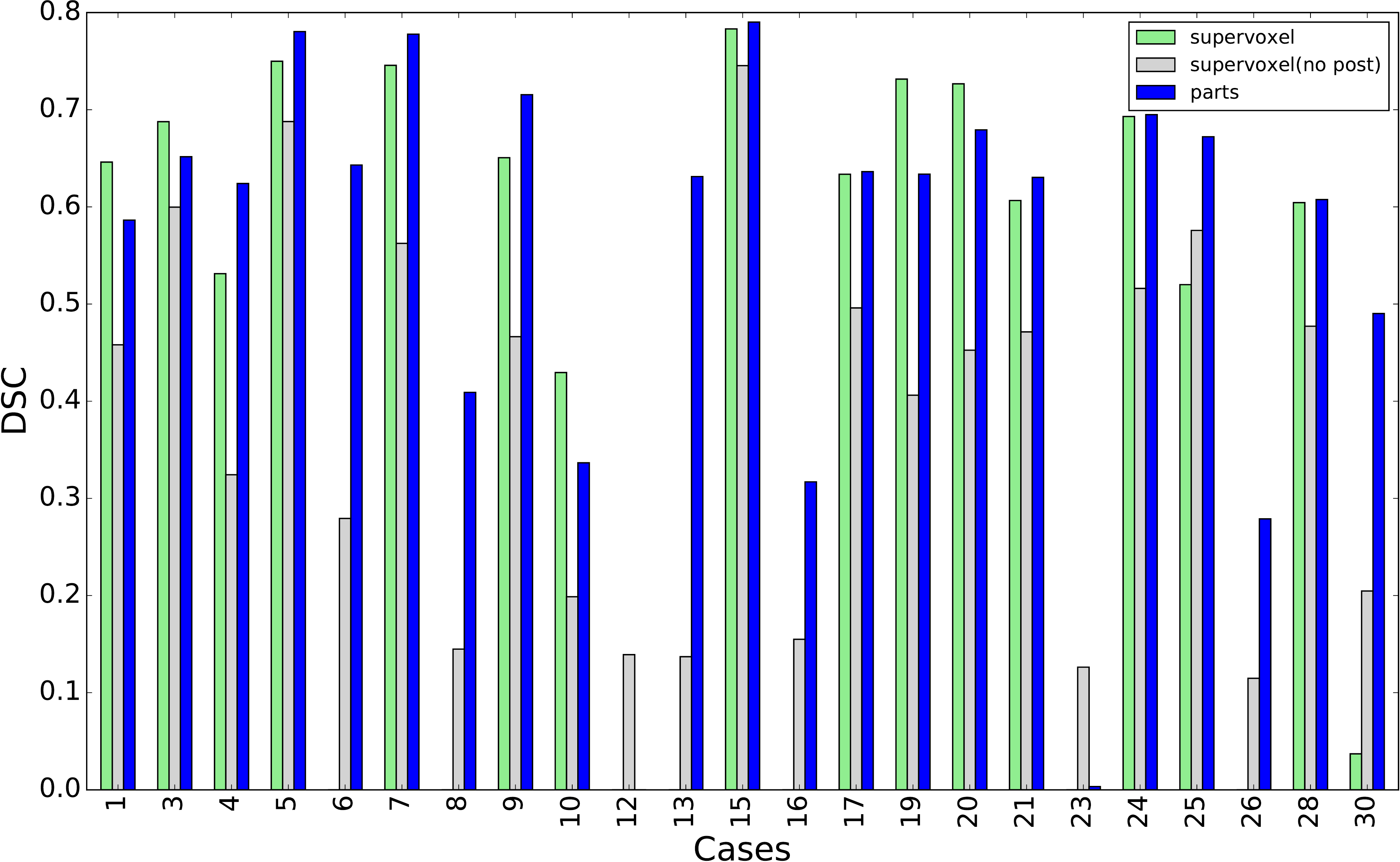}
\caption{DSC compared to the manually annotated ground truth for each cases using the perfusion-supervoxel classification (light green), the perfusion-supervoxel classification without postprocessing (grey), and the final pieces-of-parts classification (blue). The two failed cases are 12 and 23, which failed due to the limited number of female training cases in the dataset. Parameter optimisation was performed on Cases 1, 3, 4, 5, 6 (see Section \ref{sec:param})}
\label{fig:results}
\end{figure}

Figure \ref{fig:results} shows the DSC for each case\footnote{Original case numbers from the trial are used in this study. However, not all patients underwent imaging, leading to missing case numbers.} using the perfusion-supervoxel classifier ($T_s$ threshold) and the pieces-of-parts extension ($T_p$ threshold). The supervoxel approach without postprocessing ($T_p$ threshold) is also included to illustrate the connection between perfusion-supervoxels and pieces-of-parts method. While well segmented cases give similar results, poorly segmented or undetected cases are on average considerably improved using pieces-of-parts. 

\begin{figure}[h!]
\centering
\subfloat[]{
\includegraphics[width=6cm]{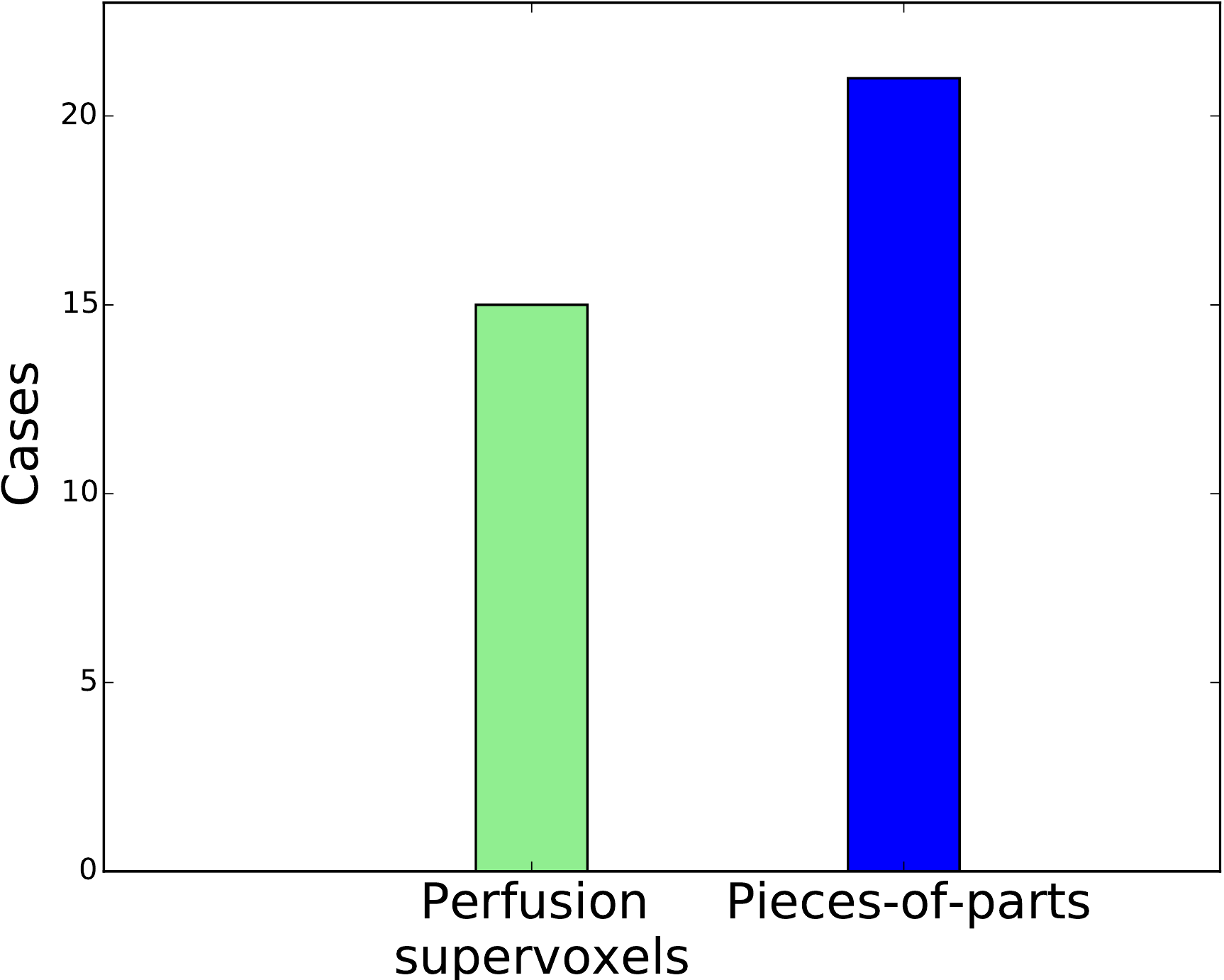}}
\subfloat[]{
\includegraphics[width=6.2cm]{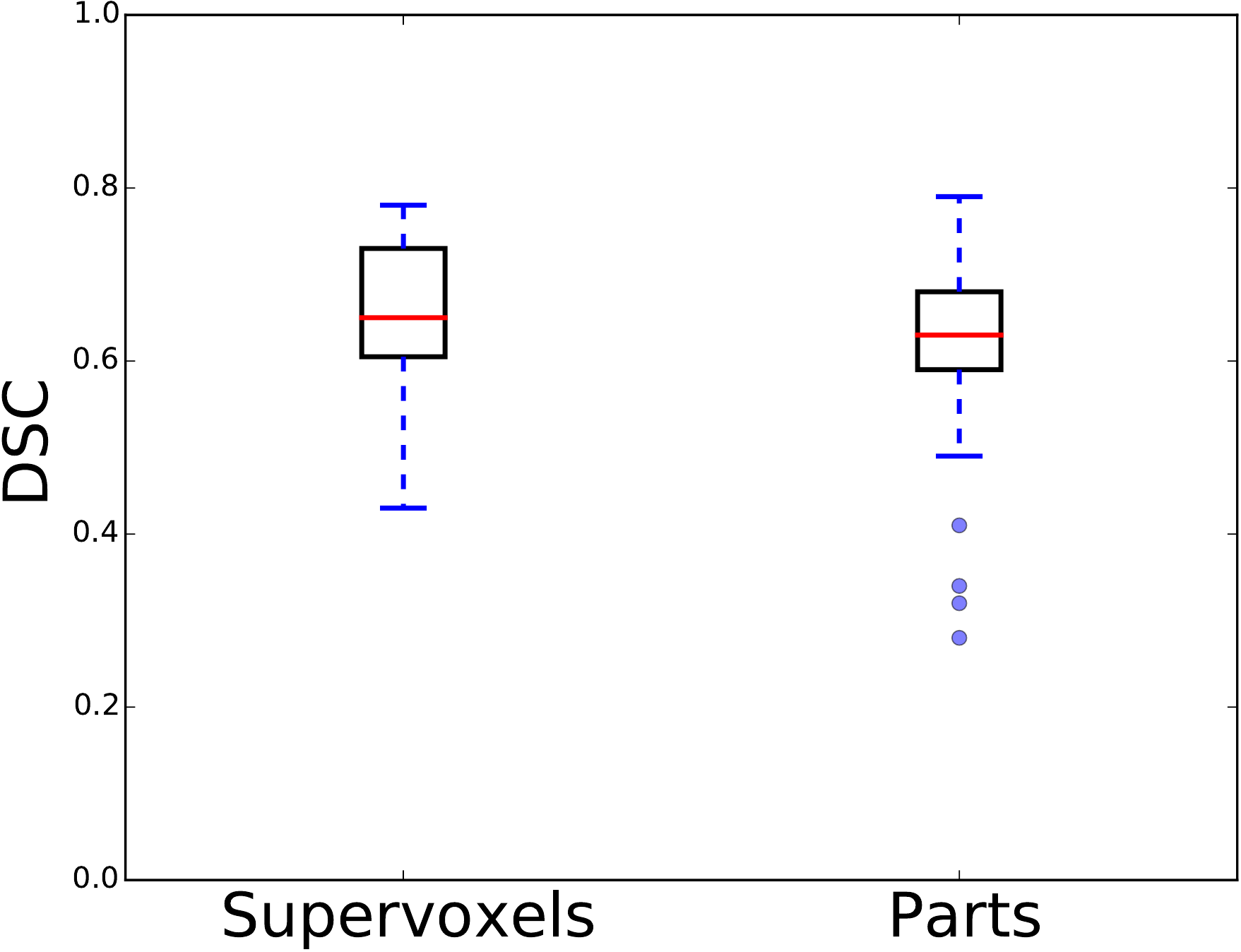}}
\caption{a) Number of cases correctly detected b) Boxplots DSC for all detected cases of each category (note that $n=15$ for the perfusion-supervoxels and $n=21$ for the pieces-of-parts method). The red bar shows the median value.} 
\label{fig:results2}
\end{figure}

The median DSC for the entire dataset was 0.60 for the initial perfusion-supervoxel classification, and slightly improved to 0.63 for the pieces-of-parts. The number of correct detections (chosen as DSC $>$20\%) is shown in Figure \ref{fig:results2} a) which increases from 15 to 21 out of 23. Box plots of the DSC of the correctly detected cases are shown in b). This clearly shows that the pieces-of-parts achieves a similar segmentation accuracy while segmenting many more cases. An influencing factor is the post-processing (Sec. \ref{sec:posproc}) of the perfusion-supervoxel method, which improves the DSC at the expense of missed cases. In this method, only the largest connected region is kept, which can lead is misclassification (such as Case 06, Figure \ref{fig:examples}) even though a response was also generated for the tumour region.

A statistical comparison was made using the Wilcoxon signed-rank test for non-parametric comparison of paired samples. Assuming significance at $p=0.05$, the pieces-of-parts method was significantly better than both the supervoxel method without post-processing ($p=0.00016$) and the full supervoxel method ($p=0.028$).

\begin{figure}[ht!]
\centering
\includegraphics[width=6cm]{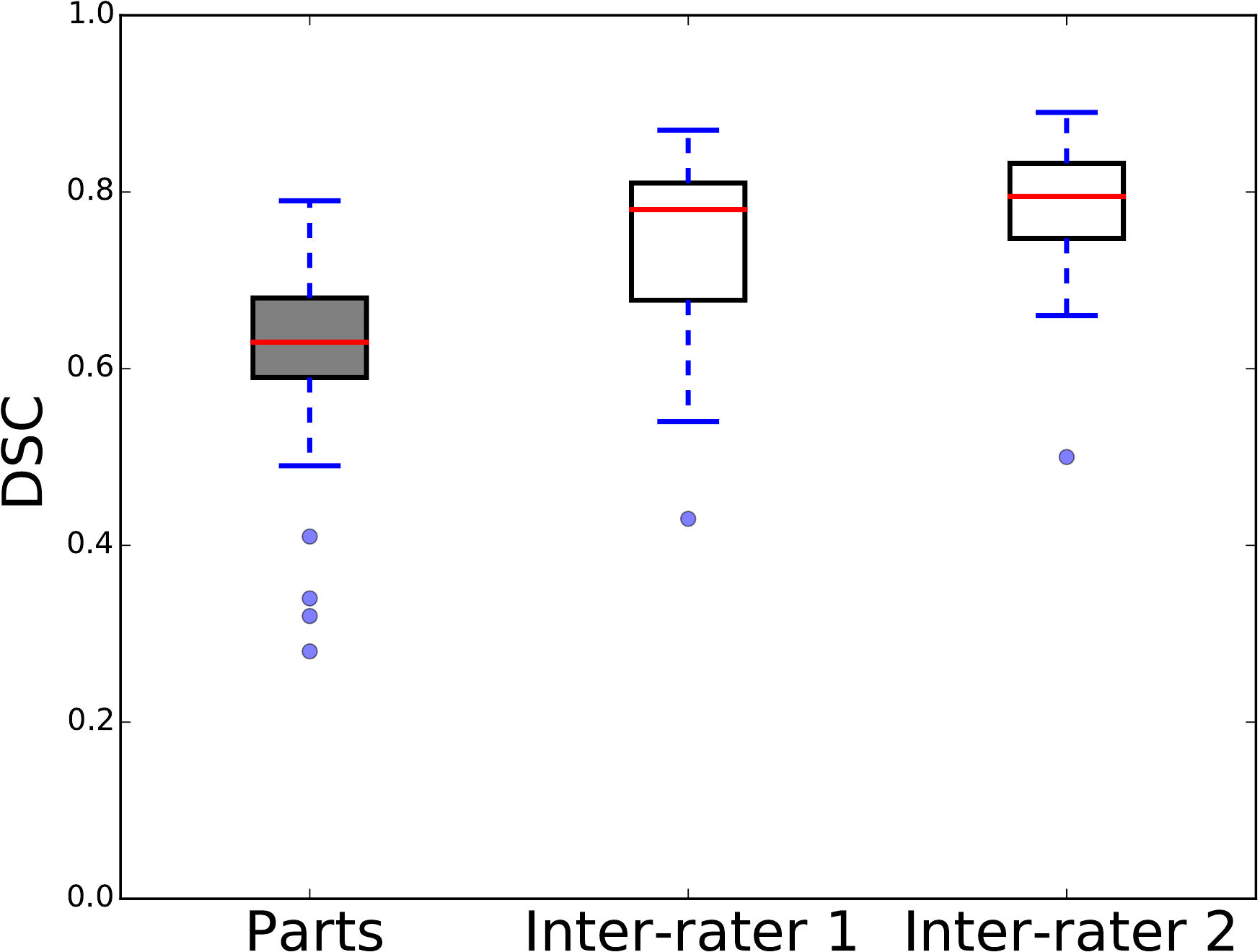}
\caption{Comparison of the automated DCE-MRI segmentation method to inter-rater variability on the T2w anatomical MRI} \label{fig:interrater}
\end{figure}

The framework presented in this paper is promising given the challenging nature of colorectal segmentation with an expert inter-rater variability of $0.73 \pm 0.13$ and $0.77 \pm 0.10$, which is essentially an upper limit on the accuracy (Section \ref{sec:mat}) and is shown in Fig. \ref{fig:interrater}. Further more, the experts both annotated on T2w images (making the experts more likely to agree), while this method detects the tumour directly from the DCE-MRI volume. Nevertheless, the automated approach is close to the inter-rater variability. 

\subsubsection{Experiment 3: Male vs female anatomy}

\begin{figure}[ht!]
\centering
\subfloat[]{
\includegraphics[width=6cm]{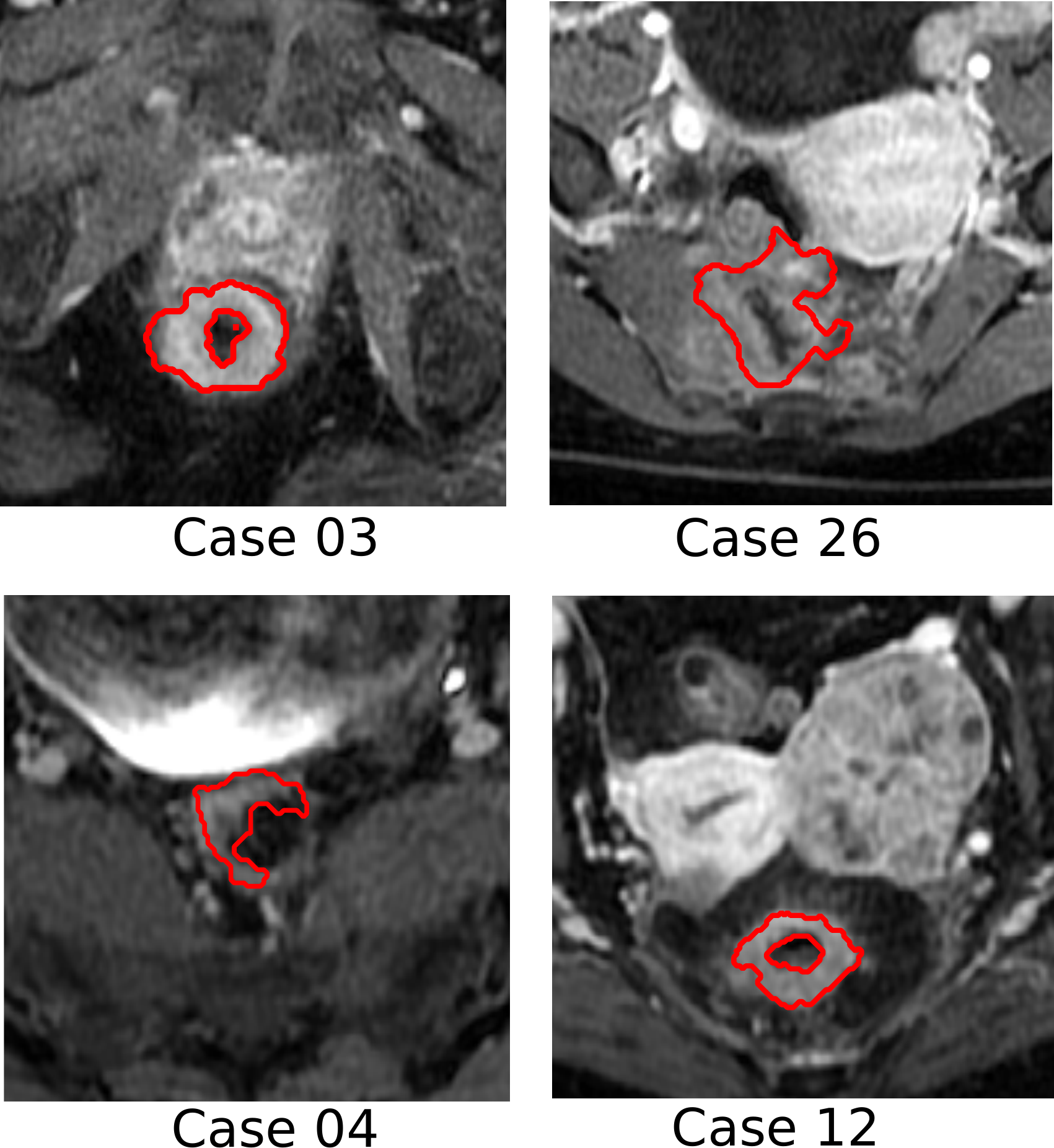}}
\subfloat[]{
\includegraphics[width=7cm]{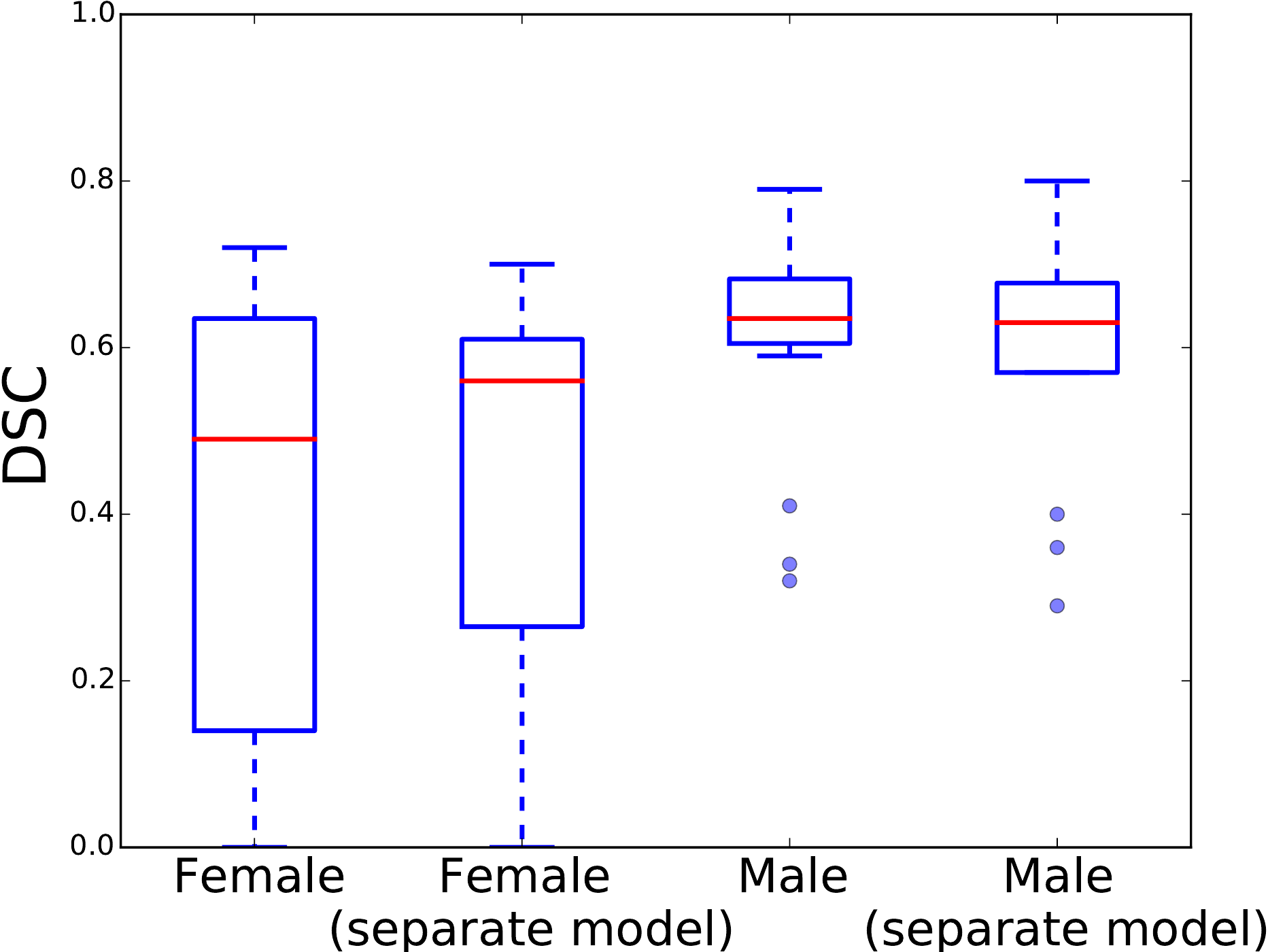}}
\caption{DSC for gender specific models. (a) Examples of male (Case 03 and Case 04, left) and female (Case 12 and Case 26, right) DCE-MRI cross sections. Case 12 has uterine fibroids present in the scan that add further complexity to the region. (b) The box plots show results for both male and female using the general model and separate gender specific models. A larger cohort is required to fully evaluate gender specific models. } \label{fig:box2}
\end{figure}

The framework we have presented is designed to be efficient for small training sets and has been found to be robust to patient variation. However, the dataset includes both male and female patients, and because of anatomical differences it would be useful to develop separate gender-specific models. Female cases are 9, 12, 17, 19, 23, 26, 30 (Fig. \ref{fig:results}), and include the poorest performing segmentations, including both failed cases (12 and 23, which are highlighted in the figure). Unfortunately, the available data contains only a small number of female patients (16 males and 7 females), making separate training challenging. Training the female cohort separately using an extended pieces-of-parts model that adds the uterus to the model shows potential, but ultimately more cases are required. Fig. \ref{fig:box2} shows the DSC for female and male patients when trained together, and when trained using a separate model. While some improvement to the female cases are obtained by separate training, male cases are unaffected. 

\subsubsection{Experiment 4: Tumour segmentation for the RHYTHM trial}
\begin{figure}[ht!]
\centering
\subfloat[]{
\includegraphics[width=6cm]{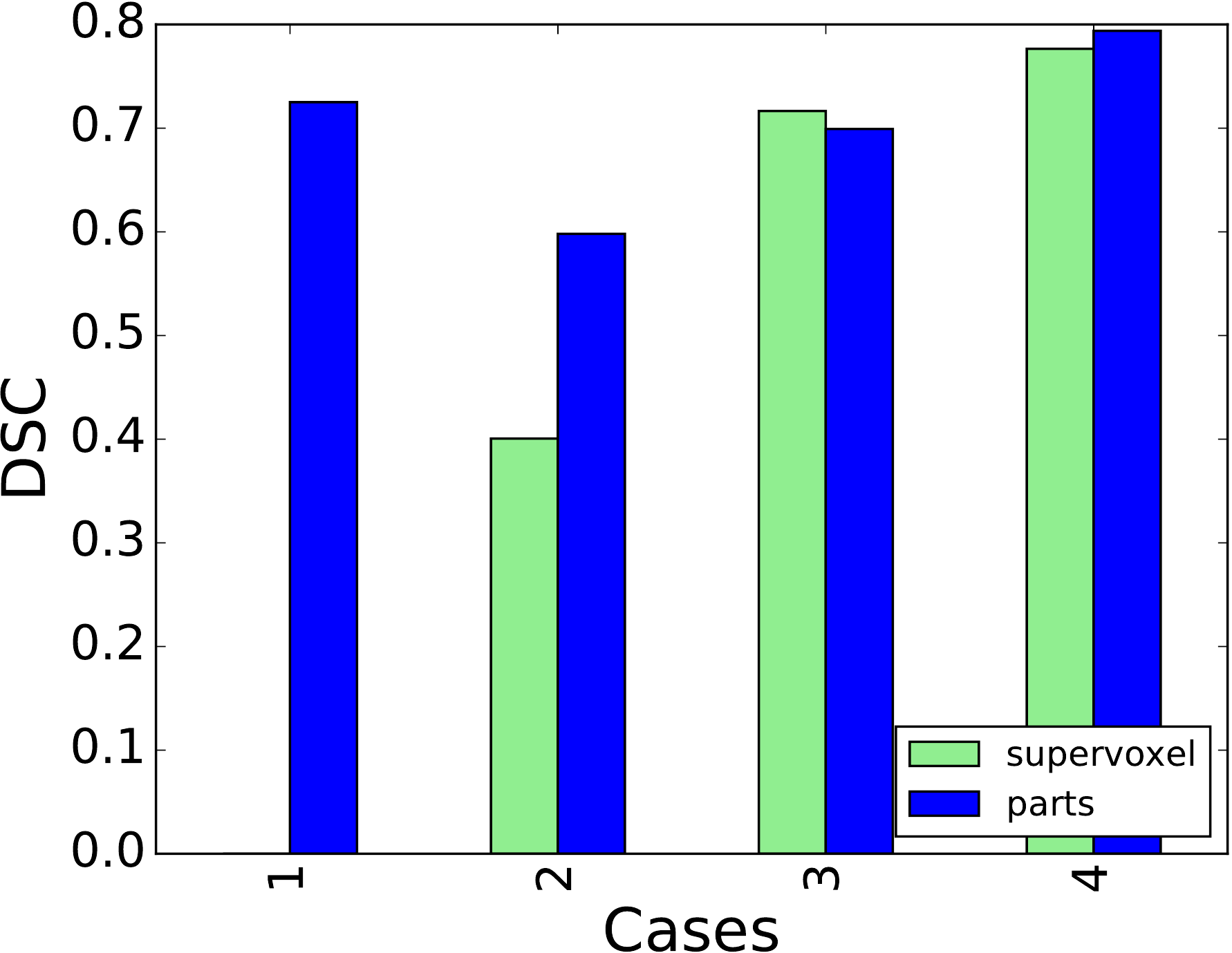}}
\subfloat[]{
\includegraphics[width=4cm]{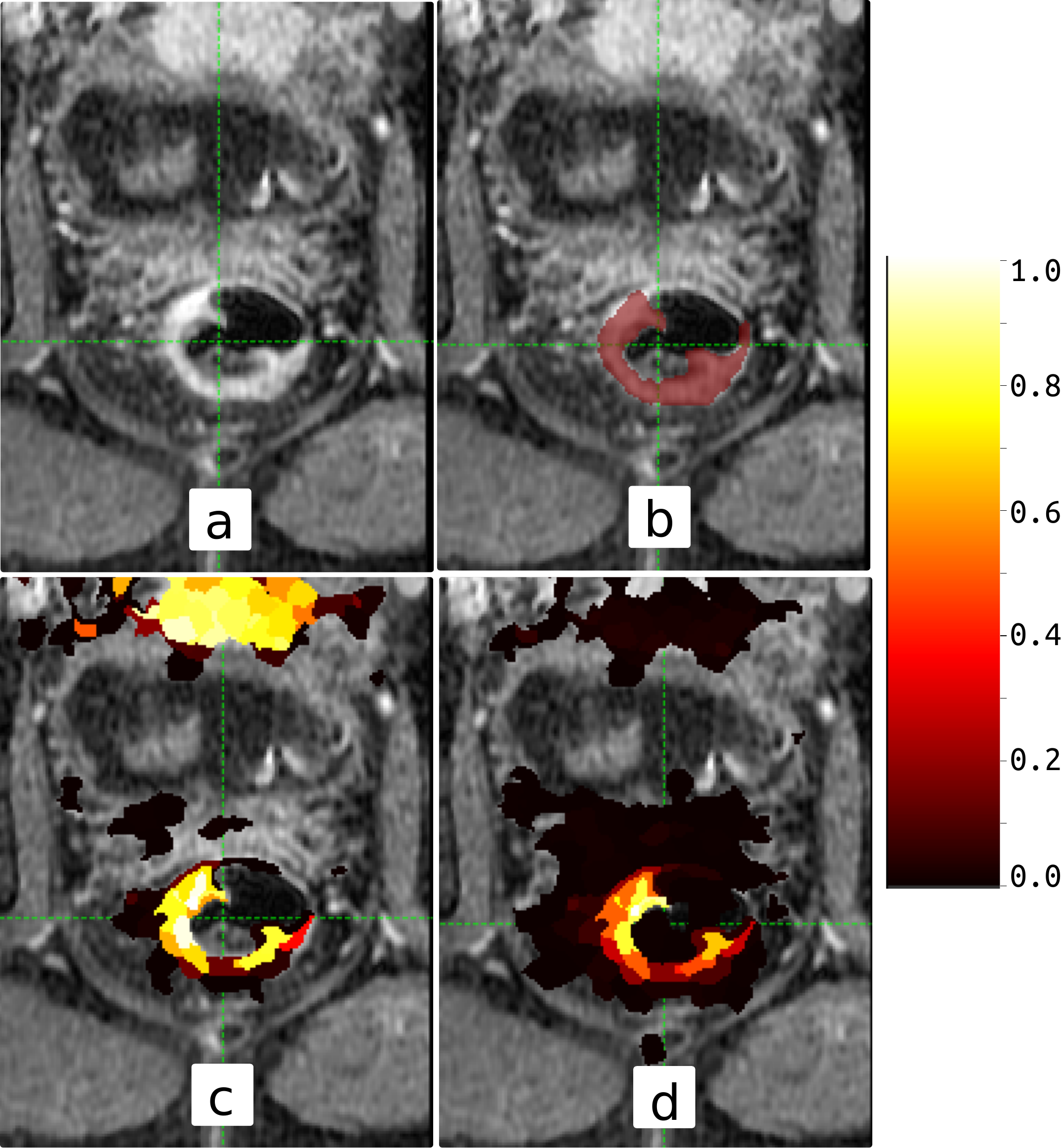}}
\caption{DSC for the RHYTHM trial and Case 1 showing a) the perfused tumour b) the manual delineation c) the perfusion-supervoxel probabilities and d) the final pieces-of-parts probabilities. Probability of zero is transparent.} \label{fig:rhb}
\end{figure}

There are only a limited number of studies available that study rectal tumours using DCE-MRI and so, given the limited data, LOOCV was used to validate this study, where each case is unseen and the method is trained on the remaining cases. However, to further demonstrate the generalisability of this method, we segmented the four DCE-MRI scans that have recently been made available to us from a new independent trial. The method was applied blindly with all training and parameter choices based on the previous study. Figure \ref{fig:rhb} shows the results. The pieces-of-parts method achieved good results with a median DSC of 0.71. Case 1 is an example of a well defined tumour that is missed by the initial supervoxel method because a larger region is also detected as tumour, but pieces-of-parts provides the global context so that the tumour is still segmented accurately. It is interesting to note that the DSC is higher for this study with 0.71 than the previous study with 0.63. This appears to be due to less patient motion in the scans.

\subsubsection{Implementation}

The framework was implemented in Python with numerically intensive steps optimised in C++ or Cython, and tested on a 6-core Intel Xeon system running Ubuntu 14.04 (x64). The mean time for a single case was $5.8 \pm 4.0$ minutes, where Table \ref{tab:time} shows the time and computational complexity for each step. These times are for a single processor, but multiprocessing was used when running a batch of cases, leading to an approximately five-fold increase in speed over the reported time. \emph{Pieces-of-parts} barely adds any additional time to the perfusion-supervoxel classifier. 

\begin{table}[ht!]
\centering
\footnotesize
\begin{tabular}{lcc}
\hline
Step & Time (s)  & Complexity\\
\hline
1. Supervoxel extraction & $160 \pm 86$s  & $O(N_v)$\\
2. Feature extraction & $162 \pm 142$ s & $O(N_s S)$ \\
3. Supervoxel classification & $3.9 \pm 1.2$s & $O(N_s)$\\
4. Pieces-of-parts:\\
\;\;\;\;4.1 Training & $3.24 \pm 0.06$s\\
\;\;\;\;4.2 Testing & $13.2 \pm 9.7$s & $O(N_s^2 p)$\\
\hline
\end{tabular}
\caption{The mean time and approximate computational complexity  of each step for a DCE-MRI volume. $N_v$ are the number of voxels, $N_s$ are the number of supervoxels ($N_s << N_v$), $S$ is mean number of voxels in a supervoxel and $p$ are the number of parts. } \label{tab:time}
\end{table}

\section{Discussion} \label{sec:discuss}

We make two key contributions in this paper: \emph{perfusion-supervoxels} and the \emph{pieces-of-parts} method. This is applied to the challenging and previously unexplored area of automated rectal tumour segmentation from DCE-MRI scans. Accurate and consistent tumour segmentation is very important in DCE-MRI as it impacts any modelling and analysis that is performed, but is a challenging and time consuming task to perform manually due to the 4D contrast-varying nature of the image. The results are very promising with 21 of the 23 cases detected correctly, an AUC of 0.97 for the voxelwise labelling of tumour and background, and a median DSC score of 0.63. A DSC of 0.71 is obtained for a second trial to illustrate the generalisability of the method.

While the methods are applied to the problem of rectal tumour segmentation from DCE-MRI, there is a broad range of other potential applications. \emph{Perfusion-supervoxels} are potentially applicable to any dynamic contrast enhanced sequence, while \emph{pieces-of-parts} can be effective for adding global constraints to superpixel/supervoxel over-segmentations. Pieces-of-parts method is suitable for images or volumes where there is an expected spatial relationship between the parts (regions of the image to be labelled) -- such as organs. The method is also better suited to problems where the parts are small relative to the entire volume so as to provide meaningful spatial constraints, and, therefore, may not be optimal for scene labelling problems, where every voxel is assigned a label. In this case a fully connected CRF would be more appropriate. In future work it would be interesting to compare the trade-offs in more detail between pieces-of-parts and a fully connected CRF for variation in the type of scene. Pieces-of-parts also does not place any shape priors on the individual parts, which is beneficial for a tumour segmentation where shape is highly variable. For segmentation of well defined objects, a shape prior could be included into the segmentation. While the method is appropriate for small datasets, the size of the dataset we use is limited and more extensive testing on a larger test dataset would be valuable. 

This method applies normalisation and signal decomposition steps in the learning algorithm but does not explicitly convert the signal to Gadolinium concentration. We found that the noise introduced using a non-linear conversion to concentration did not improve segmentation accuracy. 

As discussed previously, development of a ground truth segmentation on which to evaluate this method is challenging, due to the difficulty of delineating tumours in rectal DCE-MRI. This required delineations on T2w anatomical MRI scans, which were then registered to the DCE-MRI, and the delineations were further adjusted based on both aligned scans. This justifies the need for a automated and consistent approach to DCE-MRI rectal tumour segmentation. Even on the T2w anatomical images the mean DSC score was $0.73 \pm 0.13$ and $0.77 \pm 0.10$ between two additional experts and the main observer, which sets an upper limit on the evaluation of our method. 

There are a number of potential extensions to this framework. A challenge is the performance on female patients due the limited female patient training data and anatomical differences. Developing gender specific models shows potential but a larger training dataset is required. This method is intended for automatic segmentation but would also be appropriate in a semi-automatic system to allow clinicians to adjust the threshold and weight parameters in order to quickly modify the segmentation. 

\section*{Acknowledgements}

This research is supported by the CRUK/EPSRC Oxford Cancer Imaging Centre. 

\section*{References}
\bibliography{library.bib}

\end{document}